%% file: main.tex
\documentclass[10pt,twocolumn,letterpaper]{article}

\usepackage{iccv}
\usepackage{times}
\usepackage{epsfig}
\usepackage{graphicx}
\usepackage{amsmath}
\usepackage{amssymb}
\usepackage{url}            
\usepackage{booktabs}       
\usepackage{amsfonts}       
\usepackage{nicefrac}       
\usepackage{microtype}      
\usepackage{color}
\usepackage{xcolor}
\usepackage{xspace}
\usepackage{multirow}
\usepackage{floatrow}
\usepackage{adjustbox}
\usepackage{tabularx}
\usepackage{subcaption}
\usepackage{dsfont}

\usepackage[pagebackref=true,breaklinks=true,colorlinks,bookmarks=false]{hyperref}

\definecolor{applegreen}{rgb}{0.55, 0.71, 0.0}
\newboolean{revising}
\setboolean{revising}{true}
\ifthenelse{\boolean{revising}}
{
\newcommand{\pk}[1]{\textcolor{red}{PHILIPP: #1}}
\newcommand{\dq}[1]{\textcolor{orange}{DEQUAN: #1}}

\newcommand{\td}[1]{\textcolor{magenta}{TREVOR: #1}}
\newcommand{\lj}[1]{\textcolor{blue}{JI: #1}}

\newcommand{\question}[1]{\textcolor{red}{Question: #1}}
\newcommand{\todo}[1]{\textcolor{red}{TODO: #1}}

\newcommand{\eat}[1]{}
}{
\newcommand{\pk}[1]{} 
\newcommand{\dq}[1]{} 

\newcommand{\td}[1]{} 
\newcommand{\lj}[1]{} 

\newcommand{\question}[1]{}
\newcommand{\todo}[1]{}

\newcommand{\eat}[1]{}
}

\newfloatcommand{figurebox}{figure}[\nocapbeside][\dimexpr(\textwidth-\columnsep)/2\relax]
\newfloatcommand{tablebox}{table}[\nocapbeside][\dimexpr(\textwidth-\columnsep)/2\relax]

\newcommand{\ra}[1]{\renewcommand{\arraystretch}{#1}}

\newcommand\minisection[1]{\noindent \textbf{#1}}
\newcommand\subfix[1]{\mathtt{#1}}
\newcommand\mf[1]{\mathbf{#1}}

\DeclareMathOperator*{\argmin}{arg\,min}

\newcommand{\tabref}{Table~\ref}
\newcommand{\figref}{Figure~\ref}
\newcommand{\secref}{Section~\ref}
\newcommand{\eqnref}{Equation~\ref}
\newcommand{\apnref}{Appendix~\ref}

\iccvfinalcopy 

\def\iccvPaperID{839} 

\ificcvfinal\fi
\begin{document}

\title{Joint Monocular 3D Vehicle Detection and Tracking}

\author{
    Hou-Ning Hu$^{1}$\thanks{Work was done while Hou-Ning Hu, Qi-Zhi Cai and Ji Lin were at the Berkeley 
DeepDrive Center}, 
    Qi-Zhi Cai$^{2}$\footnotemark[1], 
    Dequan Wang$^{3}$, 
    Ji Lin$^{4}$\footnotemark[1], 
    \and
    Min Sun$^{1}$, 
    Philipp Kr\"ahenb\"uhl$^{5}$, 
    Trevor Darrell$^{3}$, 
    Fisher Yu$^{3}$
    \and
    $^{1}$National Tsing Hua University \space
    $^{2}$Sinovation Ventures AI Institute \space
    \and
    $^{3}$UC Berkeley \space
    $^{4}$MIT \space
    $^{5}$UT Austin \space
}

\maketitle


\begin{abstract}


Vehicle 3D extents and trajectories are critical cues for predicting the future location of vehicles and planning future agent ego-motion based on those predictions. 
In this paper, we propose a novel online framework for 3D vehicle detection and tracking from monocular videos. 
The framework can not only associate detections of vehicles in motion over time, but also estimate their complete 3D bounding box information from a sequence of 2D images captured on a moving platform. 
Our method leverages 3D box depth-ordering matching for robust instance association and utilizes 3D trajectory prediction for re-identification of occluded vehicles. 
We also design a motion learning module based on an LSTM for more accurate long-term motion extrapolation.
Our experiments on simulation, KITTI, and Argoverse datasets show that our 3D tracking pipeline offers robust data association and tracking.
On Argoverse, our image-based method is significantly better for tracking 3D vehicles within 30 meters than the LiDAR-centric baseline methods.



\end{abstract}

\begin{figure}[t]
	\begin{center}
		\includegraphics[width=1.0\linewidth]{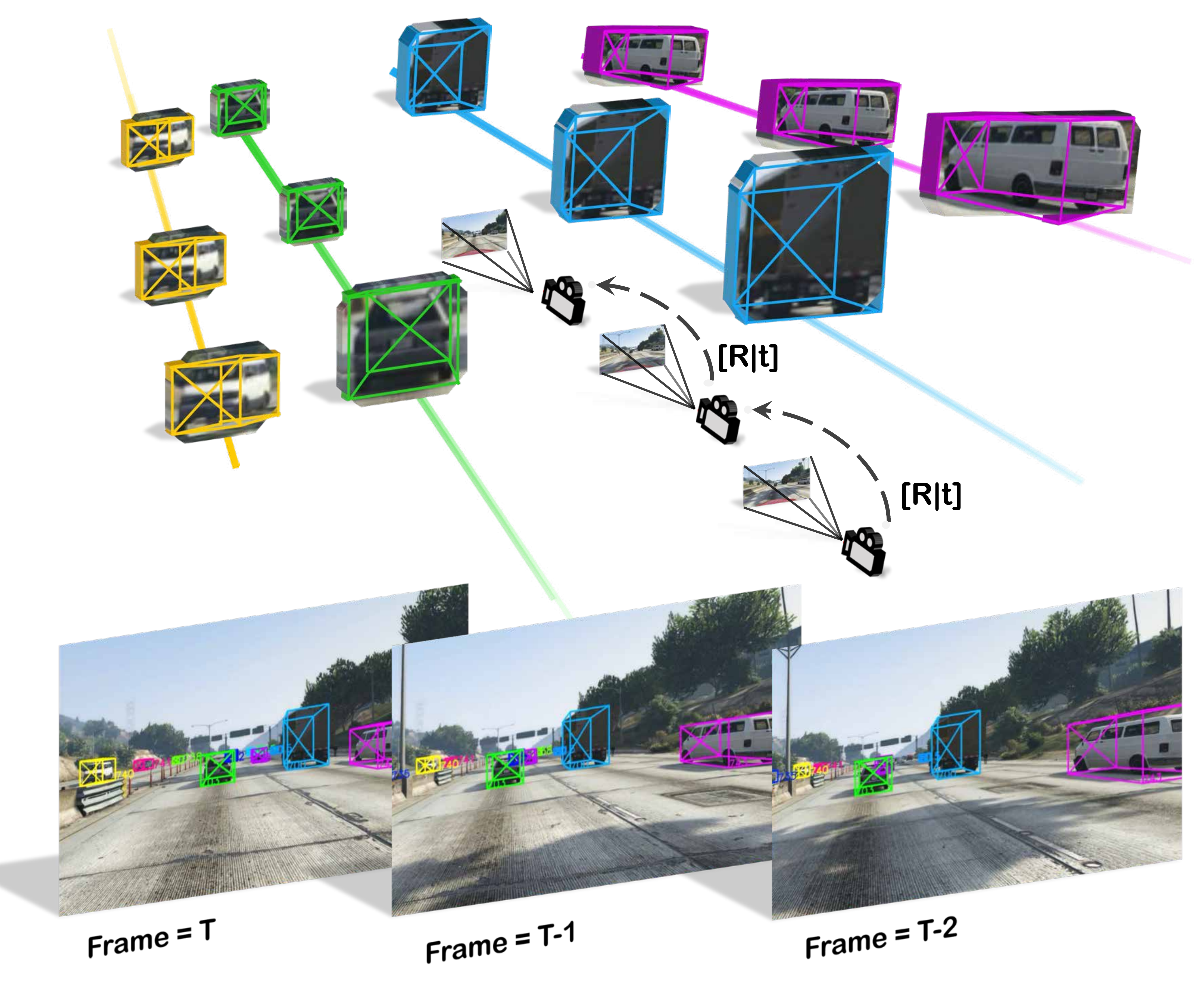}
	\end{center}
	\caption{
	Joint online detection and tracking in 3D. 
	Our dynamic 3D tracking pipeline predicts 3D bounding box association of observed vehicles in image sequences captured by a monocular camera with an ego-motion sensor.}
	\label{fig:teaser}
\end{figure}

\input{01Intro}
\input{02RelatedWorks}
\input{03Technical}
\input{04Dataset}
\input{05Experiments}

\input{06Conclusion}

\input{07Acknowledgements}

\clearpage

{\small
\bibliographystyle{ieee_fullname}
\bibliography{egbib}
}

\clearpage
\input{appendix.tex}

\end{document}

%% file: 01Intro.tex
\section{Introduction}
Autonomous driving motivates much of contemporary visual deep learning research.
However, many commercially successful approaches to autonomous driving control rely on a wide array of views and sensors, reconstructing 3D point clouds of the surroundings before inferring object trajectories in 3D.  
In contrast, human observers have no difficulty in perceiving the 3D world in both space and time from simple sequences of 2D images rather than 3D point clouds, even though human stereo vision only reaches several meters.
Recent progress in monocular object detection and scene segmentation offers the promise to make 
low-cost mobility widely available. 
In this paper, we explore architectures and datasets for developing similar capabilities using deep neural networks.

Monocular 3D detection and tracking are inherently ill-posed.
In the absence of depth measurements or strong priors, a single view does not provide enough information to estimate 3D layout of a scene accurately.
Without a good layout estimate, tracking becomes increasingly difficult, especially in the presence of large ego-motion (\eg, a turning car).
The two problems are inherently intertwined. 
Robust tracking helps 3D detection, as information along consecutive frames is integrated.
Accurate 3D detection helps to track, as ego-motion can be factored out.

In this paper, we propose an online network architecture to jointly track and detect vehicles in 3D from a series of monocular images. 
\figref{fig:teaser} provides an overview of our 3D tracking and detection task.
After detecting 2D bounding boxes of objects, we utilize both world coordinates and re-projected camera coordinates to associate instances across frames.
Notably, we leverage novel occlusion-aware association and depth-ordering matching algorithms to overcome the occlusion and reappearance problems in tracking.
Finally, we capture the movement of instances in a world coordinate system and update their 3D poses using LSTM motion estimation along a trajectory, integrating single-frame observations associated with the instance over time.

Like any deep network, our model is data hungry.
The more data we feed it, the better it performs.
However, existing datasets are either limited to static scenes~\cite{wen2015ua}, lack the required ground truth trajectories~\cite{milan2016mot16}, or are too small to train contemporary deep models~\cite{geiger2012we}.
To bridge this gap, we resort to realistic video games.
We use a new pipeline to collect large-scale 3D trajectories, from a realistic synthetic driving environment, augmented with dynamic meta-data associated with each observed scene and object.

To the best of our knowledge, we are the first to tackle the estimation of complete 3D vehicle bounding box tracking information from a monocular camera.
We jointly track the vehicles across frames based on deep features and estimate the full 3D information of the tracks including position, orientation, dimensions, and projected 3D box centers of each object. 
The depth ordering of the tracked vehicles constructs an important perceptual cue to reduce the mismatch rate. 
Our occlusion-aware data association provides a strong prior for occluded objects to alleviate the identity switch problem.
Our experiments show that 3D information improves predicted association in new frames compared to traditional 2D tracking, and that estimating 3D positions with a sequence of frames is more accurate than single-frame estimation. 



%% file: 02RelatedWorks.tex
\section{Related Works}

\minisection{Object tracking} has been explored extensively in the last decade~\cite{yilmaz2006object,salti2012adaptive,smeulders2014visual}. 
Early methods~\cite{bolme2010visual,gaidon2016virtual,kristan2015visual} track objects based on correlation filters.
Recent ConvNet-based methods typically build on pre-trained object recognition networks. 
Some generic object trackers are trained entirely online, starting from the first frame of a given video~\cite{hare2016struck,babenko2009visual,kalal2012tracking}. 
A typical tracker will sample patches near the target object which are considered as foreground and some farther patches as background. 
These patches are then used to train a foreground-background classifier. 
However, these online training methods cannot fully utilize a large amount of video data.
Held~\etal~\cite{held2016learning} proposed a regression-based method for offline training of neural networks, tracking novel objects at test-time at 100 fps. 
Siamese networks also found in use, including tracking by object verification~\cite{tao2016siamese}, tracking by correlation~\cite{bertinetto2016fully}, tracking by detection~\cite{feichtenhofer2017detect}. 
Yu~\etal~\cite{xiang2015mdptrack} enhance tracking by modeling a track-let into different states and explicitly learns an Markov Decision Process (MDP) for state transition. 
Due to the absence of 3D information, it just uses 2D location to decide whether a track-let is occluded.

All those methods only take 2D visual features into consideration, where the search space is restricted near the original position of the object. 
This works well for a static observer, but fails in a dynamic 3D environment.
Here, we further leverage 3D information to narrow down the search space, and stabilize the trajectory of target objects.

Sharma~\etal.~\cite{MOTBeyondPixels} uses 3D cues for 2D vehicle tracking. Scheidegger~\etal.~\cite{Scheidegger2018} also adds 3D kalman filter on the 3D positions to get more consistent 3D localization results. 
Because the goals are for 2D tracking, 3D box dimensions and orientation are not considered. 
Osep ~\etal.~\cite{Osep17ICRAciwt} and Li~\etal.~\cite{Li_2018_ECCV} studies 3D bounding box tracking with stereo cameras. 
Because the 3D depth can be perceived directly, the task is much easier, but in many cases such as ADAS, large-baseline stereo vision is not possible. 

\minisection{Object detection} reaped many of the benefits from the success of convolutional representation. 
There are two mainstream deep detection frameworks: 
1) two-step detectors: R-CNN~\cite{girshick2014rich}, Fast R-CNN~\cite{girshick2015fast}, and Faster R-CNN~\cite{ren2015faster}. 
2) one-step detectors: YOLO~\cite{redmon2016you}, SSD~\cite{liu2016ssd}, and YOLO9000~\cite{redmon2017yolo9000}. 

We apply Faster R-CNN, one of the most popular object detectors, as our object detection input.
The above algorithms all rely on scores of labeled images to train on.
In 3D tracking, this is no different.
The more training data we have, the better our 3D tracker performs.
Unfortunately, getting a large amount of 3D tracking supervision is hard.

\minisection{Driving datasets} have attracted a lot of attention in recent years. 
KITTI~\cite{geiger2012we}, Cityscapes~\cite{cordts2016cityscapes}, Oxford RobotCar~\cite{maddern20171}, BDD100K~\cite{yu2018bdd100k}, NuScenes~\cite{caesar2019nuscenes}, and Argoverse~\cite{Chang2019argoverse} provide well annotated ground truth for visual odometry, stereo reconstruction, optical flow, scene flow, object detection and tracking.
However, their provided 3D annotation is very limited compared to virtual datasets.
Accurate 3D annotations are challenging to obtain from humans and expensive to measure with 3D sensors like LiDAR.
Therefore these real-world datasets are typically small in scale or poorly annotated. 

To overcome this difficulty, there has been significant work on virtual driving datasets: virtual KITTI~\cite{gaidon2016virtual}, SYNTHIA~\cite{ros2016synthia}, GTA5~\cite{richter2016playing}, VIPER~\cite{richter2017playing}, CARLA~\cite{dosovitskiy2017carla}, and Free Supervision from Video Games (FSV)~\cite{pk-fsvg-2018}.
The closest dataset to ours is VIPER~\cite{richter2017playing}, which provides a suite of videos and annotations for various computer vision problems while we focus on object tracking. 
We extend FSV~\cite{pk-fsvg-2018} to include object tracking in both 2D and 3D, as well as fine-grained object attributes, control signals from driver actions.

In the next section, we describe how to generate 3D object trajectories from 2D dash-cam videos.
Considering the practical requirement of autonomous driving, we primarily focus on online tracking systems, where only the past and current frames are accessible to a tracker.

%% file: 03Technical.tex
\begin{figure*}[htb!]
		\includegraphics[width=1.0\linewidth]{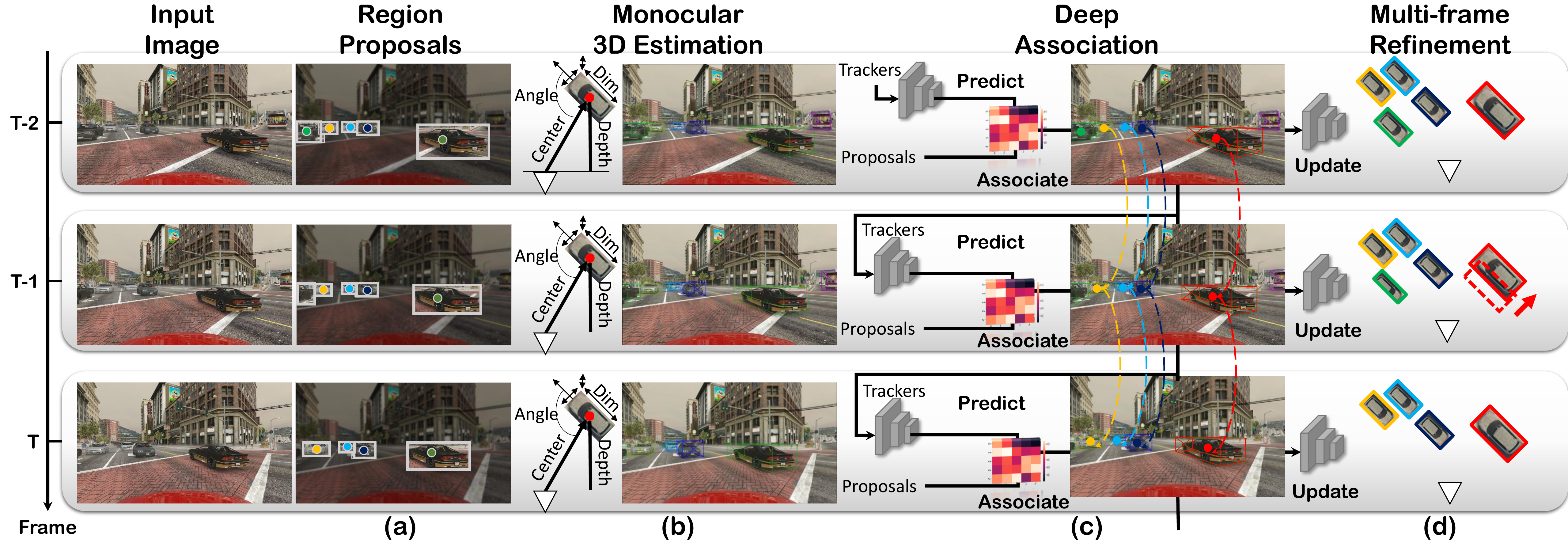}
	\caption{
	Overview of our monocular 3D tracking framework. 
	Our online approach processes monocular frames to estimate and track region of interests (RoIs) in 3D (a). 
	For each ROI, we learn 3D layout (\ie, depth, orientation, dimension, a projection of 3D center) estimation (b). 
	With 3D layout, our LSTM tracker produces robust linking across frames leveraging occlusion-aware association and depth-ordering matching (c). 
	With the help of 3D tracking, the model further refines the ability of 3D estimation by fusing object motion features of the previous frames (d).
	}
	\label{fig:overview}
\end{figure*}

\section{Joint 3D Detection and Tracking}
Our goal is to track objects and infer their precise 3D location, orientation, and dimension from a single monocular video stream and a GPS sensor.
\figref{fig:overview} shows an overview of our system.
Images are first passed through a detector network trained to generate object proposals and centers. These proposals are then fed into a layer-aggregating network which infers 3D information.
Using 3D re-projection to generate similarity metric between all trajectories and detected proposals, we leverage estimated 3D information of current trajectories to track them through time.
Our method also solves the occlusion problem in tracking with the help of occlusion-aware data association and depth-ordering matching.
Finally, we re-estimate the 3D location of objects using the LSTM through the newly matched trajectory.

\subsection{Problem Formulation}
We phrase the 3D tracking problem as a supervised learning problem. We aim to find $N$ trajectories $\{\mathbf{\tau}^1,\ldots,\mathbf{\tau}^N\}$, one for each object in a video.
Each trajectory $\mathbf{\tau}^i$ links a sequence of detected object states $\{s^{(i)}_a,s^{(i)}_{a+1},\ldots,s^{(i)}_b\}$ starting at the first visible frame $a$ and ending at the last visible frame $b$.
The state of an object at frame $a$ is given by
$s_a = (P,O,D,F,\Delta P),$
where $P$ defines the 3D world location $(x,y,z)$ of the object, and $\Delta P$ stands for its velocity $(\dot{x}, \dot{y}, \dot{z})$.
$O,D,F$ denotes for object orientation $\theta$, dimension ($l, w, h$) and appearance feature $f_{app}$, respectively. 
In addition, we reconstruct a 3D bounding box $X$ for each object, with estimated $P,O,D$ and the projection $c=({x_c}, {y_c})$ of 3D box's center in the image.
The bounding boxes enable the use of our depth-ordering matching and occlusion-aware association.
Each bounding box $X$ also forms a projected 2D box $M(X)=\{x_{\mathtt{min}},y_{\mathtt{min}},x_{\mathtt{max}},y_{\mathtt{max}}\}$ projected onto a 2D image plane using camera parameters $M=K[R|t]$.

The intrinsic parameter $K$ can be obtained from camera calibration.
The extrinsic parameter $[R|t]$ can be calculated from the commonly equipped GPS or IMU sensor.
The whole system is powered by a convolutional network pipeline trained on a considerable amount of ground truth supervision. 
Next, we discuss each component in more detail.

\subsection{Candidate Box Detection}
In the paper, we employ Faster R-CNN~\cite{ren2015faster} trained on our dataset to provide object proposals in the form of bounding boxes. 
Each object proposal (\figref{fig:overview}(a)) corresponds to a 2D bounding box $\textbf{d}=\{x_{\mathtt{min}},y_{\mathtt{min}},x_{\mathtt{max}},y_{\mathtt{max}}\}$ as well as an estimated projection of the 3D box's center $c$.
The detection results are used to locate the candidate vehicles and extract their appearance features.
However, the centers of objects' 3D bounding boxes usually do not project directly to the center of their 2D bounding boxes. 
As a result, we have to provide an estimation of the 3D box center for better accuracy.
More details about the estimation of the 3D center can be found in the supplementary material\footnote{Supplementary material of Joint Monocular 3D Vehicle Detection and Tracking can be found at https://eborboihuc.github.io/Mono-3DT/}.

\minisection{Projection of 3D box center.}
To estimate the 3D layout from single images more accurately, we extend the regression process to predict a projected 2D point of the 3D bounding box's center from an ROIpooled feature $F$ using L1 loss.
Estimating a projection of 3D center is crucial since a small gap in the image coordinate will cause a gigantic shift in 3D. 
It is worth noting that our pipeline can be used with any off-the-shelf detector and our 3D box estimation module is extendable to estimate projected 2D points even if the detector is replaced.
With the extended ROI head, the model regresses both a bounding box $\textbf{d}$ and the projection of 3D box's center $c$ from an anchor point.
ROIalign~\cite{he2017mask} is used instead of ROIpool to get the regional representation given the detected regions of interest (ROIs).
This reduces the misalignment of two-step quantization.

\subsection{3D Box Estimation}
\label{sec:3d_estimation}
We estimate complete 3D box information (\figref{fig:overview}(b)) from an ROI in the image via a feature representation of the pixels in the 2D bounding box.
The ROI feature vector $F$ is extracted from a $34$-layer DLA-up~\cite{yu2018dla} using ROIalign.
Each of the 3D information is estimated by passing the ROI features through a $3$-layer 3x3 convolution sub-network, which extends the stacked Linear layers design of Mousavian~\etal~\cite{mousavian20173d}.
We focus on 3D location estimation consisting of object center, orientation, dimension and depth, whereas~\cite{mousavian20173d} focus on object orientation and dimension from 2D boxes.
Besides, our approach integrates with 2D detection and has the potential to jointly training, while~\cite{mousavian20173d} crops the input image with pre-computed boxes.
This network is trained using ground truth depth, 3D bounding box center projection, dimension, and orientation values. A convolutional network is used to preserve spatial information.
In the case that the detector is replaced with another architecture, the center $c$ can be obtained from this sub-network.
More details of $c$ can be found at \apnref{apn:3d_center_projection}.

\minisection{3D World Location.} 
Contrasting with previous approaches, we also infer 3D location $P$ from monocular images.
The network regresses an inverse depth value $1/d$, but is trained to minimize the L1 loss of the depth value $d$ and the projected 3D location $P$.
A projected 3D location $P$ is calculated using an estimated 2D projection of the 3D object center $c$ as well as the depth $d$ and camera transformation $M$.

\minisection{Vehicle Orientation.} 
Given the coordinate distance $\hat{x} = x_c - \frac{w}{2}$ to the horizontal center of an image and the focal length $f$, we can restore the global rotation $\theta$ in the camera coordinate from $\theta_l$ with simple geometry, $\theta = (\theta_l + \arctan \frac{\hat{x}}{f})\mod 2\pi$.
Following \cite{mousavian20173d} for $\theta_l$ estimation, we first classify the angle into two bins and then regress the residual relative to the bin center using Smooth L1 loss.

\minisection{Vehicle Dimension.} 
In driving scenarios, the high variance of the distribution of the dimensions of different categories of vehicles (e.g., car, bus) results in difficulty classifying various vehicles using unimodal object proposals. 
Therefore, we regress a dimension $D$ to the ground truth dimension over the object feature representation using L1 loss.

The estimation of an object's 3D properties provides us with an observation for its location $P$ with  orientation $\theta$, dimension $D$ and 2D projection of its 3D center $c$.
For any new tracklet, the network is trained to predict monocular object state $s$ of the object by leveraging ROI features.
For any previously tracked object, the following association network is able to learn a mixture of a multi-view monocular 3D estimates by merging the object state from last visible frames and the current frame.
First, we need to generate such a 3D trajectory for each tracked object in world coordinates.

\subsection{Data Association and Tracking}
\label{sec:3d_tracking}
Given a set of tracks $\{\mathbf{\tau}^J,\ldots,\mathbf{\tau}^K\}$ at frame $a$ where $1 \leq J \leq K \leq M$ from $M$ trajectories, our goal is to associate each track with a candidate detection, spawn new tracks, or end a track (\figref{fig:overview}(c)) in an online fashion.

We solve the data association problem by using a weighted bipartite matching algorithm. Affinities between tracks and new detections are calculated from two criteria: overlap between projections of current trajectories forward in time and bounding boxes candidates; and the similarity of the deep representation of the appearances of new and existing object detections.
Each trajectory is projected forward in time using the estimated velocity of an object and camera ego-motion.
Here, we assume that ego-motion is given by a sensor, like GPS, an accelerometer, gyro and/or IMU.

We define an affinity matrix $\mathbf{A}(\mathbf{\tau}_a, \mathbf{s_a})$ between the information of an existing track $\mathbf{\tau}_a$ and a new candidate $\mathbf{s_a}$ as a joint probability of appearance and location correlation.
\begin{equation}
\begin{split}
    \mathbf{A}_{\subfix{deep}}(\mathbf{\tau}_a, \mathbf{s}_a) &= \exp(-||F_{\mathbf{\tau}_a}, F_{\mathbf{s}_a}||_1)
\label{eqn:affinity_deep}
\end{split}
\end{equation}
\begin{equation}
\begin{split}
    \mathbf{A}_{\subfix{2D}}(\mathbf{\tau}_a, \mathbf{s}_a) &= \frac{\mathbf{d}_{\mathbf{\tau}_a} \cap \mathbf{d}_{\mathbf{s}_a}}
    {\mathbf{d}_{\mathbf{\tau}_a} \cup \mathbf{d}_a}
\label{eqn:affinity_2d}
\end{split}
\end{equation}
\begin{equation}
\begin{split}
    \mathbf{A}_{\subfix{3D}}(\mathbf{\tau}_a, \mathbf{s}_a) &= \frac{\mf{M}(X_{\mathbf{\tau}_a}) \cap \mf{M}(X_{\mathbf{s}_a})}
    {\mf{M}(X_{\mathbf{\tau}_a}) \cup \mf{M}(X_{\mathbf{s}_a})},
\label{eqn:affinity_3d}
\end{split}
\end{equation}
where $F_{\mathbf{\tau}_a}$, $F_{\mathbf{s}_a}$ are the concatenation of appearance feature $f_{app}$, dimension $D$, center $c$, orientation $\theta$ and depth $d$. 
$X_{\mathbf{\tau}_a}$ and $X_{\mathbf{s}_a}$ are the tracked and predicted 3D bounding boxes, $\mf{M}$ is the projection matrix casting the bounding box to image coordinates, $\mathbf{A}_{\subfix{2D}}$ and $\mathbf{A}_{\subfix{3D}}$ is the Intersection of Union (IoU).
\begin{equation}
\begin{split}
    \mathbf{A}(\mathbf{\tau}_a, \mathbf{s}_a) &= w_{\subfix{deep}}  \mathbf{A}_{\subfix{app}}(\mathbf{\tau}_a, \mathbf{s}_a) + w_{\subfix{2D}}  \mathbf{A}_{\subfix{2D}}(\mathbf{\tau}_a, \mathbf{s}_a) \\ &+ w_{\subfix{3D}}  \mathbf{A}_{\subfix{3D}}(\mathbf{\tau}_a, \mathbf{s}_a)
\label{eqn:affinity_sum}
\end{split}
\end{equation}
$w_{\subfix{deep}},w_{\subfix{2D}}, w_{\subfix{3D}}$ are the weights of appearance, 2D overlap, and 3D overlap. 
We utilize a mixture of those factors as the affinity across frames, similar to the design of POI~\cite{yu2016poi}.

Comparing to 2D tracking, 3D-oriented tracking is more robust to ego-motion, visual occlusion, overlapping, and re-appearances. 
When a target is temporally occluded, the corresponding 3D motion estimator can roll-out for a period of time and relocate 2D location at each new point in time via the camera coordinate transformation.

\begin{figure}[t]
	\begin{center}
		\includegraphics[width=1.0\linewidth]{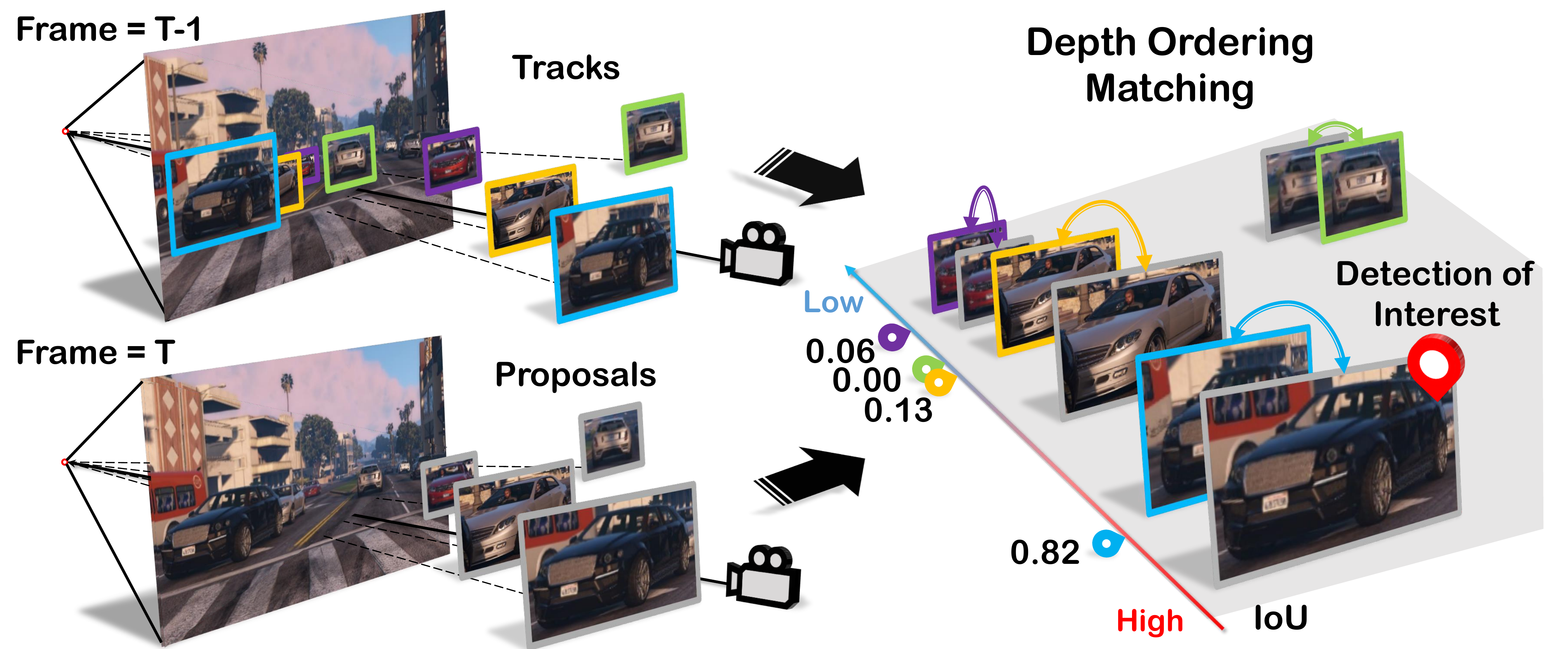}
	\end{center}
	\caption{Illustration of depth-ordering matching. Given the tracklets and detections, we sort them into a list by depth order. 
	For each detection of interest (DOI), we calculate the IOU between DOI and non-occluded regions of each tracklet.
	The depth order naturally provides higher probabilities to tracklets near the DOI.
	}
	\label{fig:ordering}
\end{figure}

\minisection{Depth-Ordering Matching.}
We introduce instance depth ordering for assigning a detection to neighbor tracklets, which models the strong prior of relative depth ordering found in human perception.
For each detection of interest (DOI), we consider potential associated tracklets in order of their depths.
From the view of each DOI, we obtain the IOU of two non-occluded overlapping map from both ascending and descending ordering.
To cancel out the ordering ambiguity of a distant tracklet, we filter out those tracklets with a larger distance to a DOI than a possible matching length.
So \eqnref{eqn:affinity_3d} becomes 
\begin{equation}
\begin{split}
    \mathbf{A}_{\subfix{3D}}(\mathbf{\tau}_a, \mathbf{s}_a) &= \mathds{1}\times\frac{\phi(\mf{M}(X_{\mathbf{\tau}_a})) \cap \mf{M}(X_{\mathbf{s}_a})}
    {\phi(\mf{M}(X_{\mathbf{\tau}_a})) \cup \mf{M}(X_{\mathbf{s}_a})},
\label{eqn:affinity_3d_update}
\end{split}
\end{equation}
where $\mathds{1}$ denotes if the tracklets is kept after depth filtering, and the overlapping function 
$$\phi(\cdot)=\argmin_x \{x | \mathtt{ord}(x) < \mathtt{ord}(x_0) \forall x_0 \in \mf{M}(X_{\mathbf{\tau}_a})) \} $$ captures pixels of non-occluded tracklets region with the nearest depth order.
It naturally provides higher probabilities of linking neighbor tracklets than those layers away.
In this way, we obtain the data association problem of moving objects with the help of 3D trajectories in world coordinates. 
\figref{fig:ordering} depicts the pipeline of depth ordering.
Finally, we solve data association using the Kuhn-Munkres algorithm.

\begin{figure}[t]
	\begin{center}
		\includegraphics[width=1.0\linewidth]{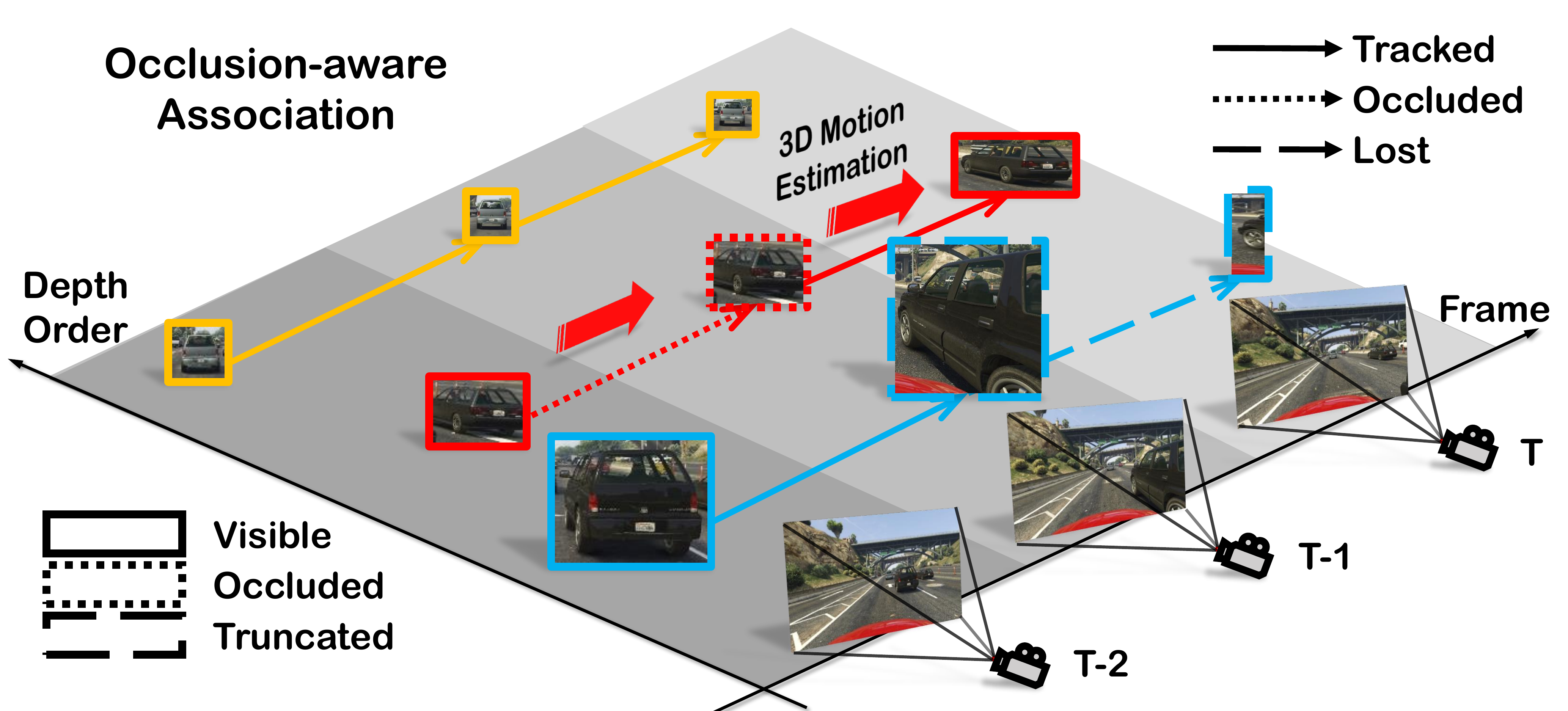}
	\end{center}
	\caption{Illustration of Occlusion-aware association. 
	A tracked tracklet (yellow) is visible all the time, while a tracklet (red) is occluded by another (blue) at frame $T-1$. 
	During occlusion, the tracklet does not update state but keep inference motion until reappearance. 
	For a truncated or disappear tracklet (blue at frame $T$), we left it as lost.
	}
	\label{fig:occlusion}
\end{figure}

\minisection{Occlusion-aware Data Association.} 
Similar to previous state-of-the-art methods~\cite{Wojke2017deepsort,xiang2015mdptrack,Sadeghian2017untrackable}, we model the lifespan of a tracker into four major subspaces in MDP state space: $\{\mathtt{birth}, \mathtt{tracked}, \mathtt{lost}, \mathtt{death}\}$.
For each new set of detections, the tracker is updated using pairs with the highest affinities score (\eqnref{eqn:affinity_sum}).
Each unmatched detection spawns a new tracklet; however, an unmatched tracklet is not immediately terminated, as tracklets can naturally disappear in occluded region and reappear later.
We address the dynamic object inter-occlusion problem by separating a new state called ``$\mathtt{occluded}$" from a $\mathtt{lost}$ state.
An object is considered $\mathtt{occluded}$ when covered by another object in the front with over $70\%$ overlap.
An $\mathtt{occluded}$ tracklet will not update its lifespan or its feature representation until it is clear from occlusion, but we still predict its 3D location using the estimated motion.
\figref{fig:occlusion} illustrates how the occlusion-aware association works.
More details of data association can be found at \apnref{apn:data_association}.

In the next subsection, we show how to estimate that distance leveraging the associated tracklet and bounding box using a deep network.

\subsection{Motion Model}
\label{sec:lstm_motion}

\minisection{Deep Motion Estimation and Update.}
To exploit the temporal consistency of certain vehicles, we associate the information across frames by using two LSTMs.
We embed a 3D location $P$ to a $64$-dim location feature and use $128$-dim hidden state LSTMs to keep track of a 3D location from the $64$-dim output feature.

Prediction LSTM (P-LSTM) models dynamic object location in 3D coordinates by predicting object velocity from previously updated velocities $\dot{P}_{T-n:T-1}$ and the previous location $\overline{P}_T-1$.
We use previous $n=5$ frames of vehicle velocity to model object motion and acceleration from the trajectory.
Given the current expected location of the object from 3D estimation module, the Updating LSTM (U-LSTM) considers both current $\hat{P}_{T}$ and previously predicted location $\tilde{P}_{T-1}$ to update the location and velocity (\figref{fig:overview}(c)).

Modeling motion in 3D world coordinates naturally cancels out adverse effects of ego-motion, allowing our model to handle missed and occluded objects.
The LSTMs continue to update the object state 
$$\mathbf{s}_a^{(i)}=\mathbf{s}_{a-1}^{(i)}+ \alpha (\mathbf{s}_a^* - \mathbf{s}_{a-1}^{(i)})$$
using the observation of matched detection state $\mathbf{s}_a^*$ with an updating ratio $\alpha=1-\mathbf{A}_{deep}(\mathbf{\tau}_a^i, \mathbf{s}_a^*)$, while assuming a linear velocity model if there is no matched bounding box.
Therefore, we model 3D motion (\figref{fig:overview}(d)) in world coordinates allowing occluded tracklet to move along motion plausible paths while managing the birth and death of moving objects.
More details can be found at \apnref{apn:motion_model}.

Concretely, our pipeline consists of a single-frame monocular 3D object detection model for object-level pose inference and recurrent neural networks for inter-frame object association and matching.
We extend the region processing to include 3D estimation by employing multi-head modules for each object instance.
We introduced occlusion-aware association to solve inter-object occlusion problem.
For tracklet matching, depth ordering lowers mismatch rate by filtering out distant candidates from a target.
The LSTM motion estimator updates the velocity and states of each object independent of camera movement or interactions with other objects.
The final pipeline produces accurate and dense object trajectories in 3D world coordinate system.

%% file: 04Dataset.tex
\section{3D Vehicle Tracking Simulation Dataset}

It is laborious and expensive to annotate a large-scale 3D bounding box image dataset even in the presence of LiDAR data, although it is much easier to label 2D bounding boxes on tens of thousands of videos~\cite{yu2018bdd100k}. 
Therefore, no such dataset collected from real sensors is available to the research community. 
To resolve the data problem, we turn to driving simulation to obtain accurate 3D bounding box annotations at no cost of human efforts. 
Our data collection and annotation pipeline extend the previous works like VIPER~\cite{richter2017playing} and FSV~\cite{pk-fsvg-2018}, especially in terms of linking identities across frames.
Details on the thorough comparison to prior data collection efforts, and dataset statistics can be found in the \apnref{apn:data_stats}.

Our simulation is based on \textit{Grand Theft Auto V}, a modern game that simulates a functioning city and its surroundings in a photo-realistic three-dimensional world.
To associate object instances across frames, we utilize in-game API to capture global instance id and corresponding 3D annotations directly. 
In contrast, VIPER leverages a weighted matching algorithm based on a heuristic distance function, which can lead to inconsistencies.
It should be noted that our pipeline is real-time, providing the potential of large-scale data collection, while VIPER requires expensive off-line processing.

%% file: 05Experiments.tex
\section{Experiments}

We evaluate our 3D detection and tracking pipeline on Argoverse Tracking benchmark~\cite{Chang2019argoverse}, KITTI MOT benchmark~\cite{geiger2012we} and our large-scale dataset, featuring real-world driving scenes and a wide variety of road conditions in a diverse virtual environment, respectively.

\subsection{Training and Evaluation}

\minisection{Dataset.} 
Our GTA raw data is recorded at $12$ FPS, which is helpful for temporal aggregation.
With the goal of autonomous driving in mind, we focus on vehicles closer than $150m$, and also filtered out the bounding boxes whose areas are smaller than $256$ pixels.
The dataset is then split into train, validation and test set with ratio $10:1:4$.
The KITTI Tracking benchmark provides real-world driving scenario.
Our 3D tracking pipeline train on the whole training set and evaluate the performance on the public testing benchmark.
The Argoverse Tracking benchmark offers novel 360 degree driving dataset.
We train on the training set and evaluate the performance on the validation benchmark since the evaluation server is not available upon the time of submission.

\minisection{Training Procedure.} We train our 3D estimation network (\secref{sec:3d_estimation}) on each training set, separately.
3D estimation network produces feature maps as the input of ROIalign~\cite{he2017mask}.
The LSTM motion module (\secref{sec:lstm_motion}) is trained on the same set with a sequence of $10$ images per batch.
For GTA, all the parameters are searched using validation set with detection bounding boxes from Faster R-CNN.
The training is conducted for $100$ epochs using Adam optimizer with an initial learning rate $10^{-3}$, momentum $0.9$, and weight decay $10^{-4}$. 
Each GPU has $5$ images and each image with no more than $300$ candidate objects before NMS.
More training details can be found in \apnref{apn:training_detail}.

\minisection{3D Estimation.} We adapt depth evaluation metrics \cite{eigen2014depth} from image-level to object-level, leveraging both error and accuracy metrics. Error metrics include absolute relative difference (Abs Rel), squared relative difference (Sq Rel), root mean square error (RMSE) and RMSE log. Accuracy metrics are percents of $y_i$ that $\mathtt{max}(\frac{y_i}{y_i^*}, \frac{y_i^*}{y_i})<\delta$ where $\delta=1.25, 1.25^2, 1.25^3$.
Following the setting of KITTI~\cite{geiger2012we}, we use orientation score (OS) for orientation evaluation.

We propose two metrics for evaluating estimated object dimension and 3D projected center position.
A Dimension Score ($DS$) measures how close an object volume estimation to a ground truth.
$DS$ is defined as 
$DS = \min(\frac{V_{\mathtt{pred}}}{V_{\mathtt{gt}}}, \frac{V_{\mathtt{gt}}}{V_{\mathtt{pred}}})$
with an upper bound $1$, where $V$ is the volume of a 3D box by multiplying its dimension $l * w * h$.
A Center Score ($CS$) measures distance of a projected 3D center and a ground truth.
$CS$ is calculated by 
$CS = (1+\cos(a_{\mathtt{gt}} - a_{\mathtt{pd}}))/2,$
with a upper bound $1$, where $a$ depicts an angular distance $((x_{\mathtt{gt}}-x_\mathtt{pd})/w_\mathtt{pd}, (y_\mathtt{gt}-y_\mathtt{pd})/h_\mathtt{pd}),$ weighted by corresponding box width and height in the image coordinates.

\minisection{Object Tracking.} 
We follow the metrics of CLEAR~\cite{bernardin2008evaluating}, including 
multiple object tracking accuracy (MOTA),
multiple object tracking precision (MOTP),
miss-match (MM),
false positive (FP), 
and false negative (FN), etc.

\minisection{Overall Evaluation.}
We evaluated the 3D IoU mAP of 3D layout estimation with refined depth estimation of different tracking methods.
The metric reflects the conjunction of all 3D components, dimension, rotation, and depth.

\subsection{Results}

\begin{table}[tpb]
    \small
    \centering
    \adjustbox{width=\linewidth}{
        \begin{tabular}{ccc|cccc}
        \toprule
        Motion & Deep & Order & MOTA$\uparrow$ & FN$\downarrow$ & MM$\downarrow$  & Ratio (\%)$\uparrow$ \\ 
        \midrule
        KF3D & - & - & 69.616 & 21.298 & 17480 & 0\\
        KF3D & - & \checkmark & 70.061 & 21.319 & \textbf{16042} & 8.222 \\
        \midrule
        KF3D & \checkmark & - & 69.843 & 18.547 & 27639 & 0 \\
        KF3D & \checkmark & \checkmark & 70.126 & 18.177 & 25860 & 6.434 \\
        LSTM & \checkmark & \checkmark & \textbf{70.439} & \textbf{18.094} & 23863 & 13.662 \\
        \bottomrule
    \end{tabular}
    }
    \caption{
    Ablation study of tracking performance with different methods in GTA dataset.
    Motion column shows which motion model being used.
    KF stands for Kalman Filter.
    Ratio is the relative improvement in reducing the number of mismatch pairs.
    Using deep feature in correlation can reduce the false negative (FN) rate. 
    Using depth-order matching and occlusion-aware association filter out relatively $6-8\%$ ($\frac{(17480-16042)}{16042}\times 100\%$) mismatching trajectories. 
    LSTM modeling dynamic motion benefits 3D IoU AP in \tabref{tab:tracking_iou}.
    }
    \vspace{-2mm}
    \label{tab:gta_tracking}
\end{table}

\begin{table}[tbp]
\centering
\small
\adjustbox{width=\linewidth}{
\begin{tabular}{c|ccc|ccc|ccc}
    \toprule
    \multirow{2}{*}{Method} & \multicolumn{3}{c|}{Easy} & \multicolumn{3}{c|}{Medium} & \multicolumn{3}{c}{Hard} \\
    & $AP_\mathtt{3d}^{70}$ & $AP_\mathtt{3d}^{50}$ & $AP_\mathtt{3d}^{25}$ & $AP_\mathtt{3d}^{70}$  & $AP_\mathtt{3d}^{50}$  & $AP_\mathtt{3d}^{25}$ & $AP_\mathtt{3d}^{70}$ & $AP_\mathtt{3d}^{50}$ & $AP_\mathtt{3d}^{25}$ \\
    \midrule
    None & 6.13   & 35.12  & 69.52  & 4.93    & 24.25   & 53.26  & 4.05   & 17.26  & 41.33  \\
    KF3D & 6.14   & 35.15  & 69.56  & 4.94    & 24.27   & 53.29  & 4.06   & 17.27  & 41.42  \\
    LSTM & \textbf{7.89}   & \textbf{36.37}  & \textbf{73.39}  & \textbf{5.25} & \textbf{26.18}  & \textbf{53.61}  & \textbf{4.46}  & \textbf{17.62}  & \textbf{41.96}  \\
    \bottomrule
\end{tabular}
}
\caption{
Comparison of tracking performance on 3D IoU AP $25$, $50$, $70$ in GTA dataset. 
Noted that KF3D slightly improves the AP performance compare to single-frame estimation (None), while LSTM grants a clear margin.
The difference comes from how we model object motion in the 3D coordinate.
Instead of using Kalman filter smoothing between prediction and observation, we directly model vehicle movement using LSTMs.
}
\vspace{-2mm}
\label{tab:tracking_iou}
\end{table}

\begin{table}[tpb]
\small
\vspace{-2mm}
\caption{Importance of using projection of 3D bounding box center estimation on KITTI training set.
We evaluate our proposed model using different center inputs $c$ to reveal the importance of estimating projection of a 3D center.
The reduction of ID Switch (IDS), track fragmentation (FRAG), and the increase of MOTA suggest that the projection of a 3D center benefits our tracking pipeline over the 2D center.
}
\centering
\adjustbox{width=\linewidth}{
    \begin{tabular}{c|cccc|cc}
        \toprule
         & IDS $\downarrow$ & FRAG $\downarrow$ & FP $\downarrow$ & FN $\downarrow$ & MOTA $\uparrow$ & MOTP $\uparrow$ \\ 
        \midrule
        2D Center & 315 & 497 & 401 & 1435 & 91.06 & 90.36 \\
        3D Center & \textbf{190} & \textbf{374} &  \textbf{401} &  \textbf{1435} &  \textbf{91.58} &  \textbf{90.36} \\
        \bottomrule
    \end{tabular}
}
\vspace{-2mm}
\label{tab:projection_3d_center}
\end{table}


\begin{table*}[tpb]
\small
\caption{Performance of 3D box estimation. 
The evaluation demonstrates the effectiveness of our model from each separate metrics. 
The models are trained and tested on the same type of dataset, either GTA or KITTI.
With different amounts of training data in our GTA dataset, the results suggest that large data capacity benefits the performance of a data-hungry network.}
\centering
\adjustbox{width=\linewidth}{
    \begin{tabular}{lr|rrrr|rrr|r|r|r}
    \toprule
    \multirow{2}{*}{Dataset} & \multirow{2}{*}{Amount} & \multicolumn{4}{c|}{Depth Error Metric} & \multicolumn{3}{c|}{Depth Accuracy Metric} & \multicolumn{1}{c|}{Orientation} & \multicolumn{1}{c|}{Dimension} &
    \multicolumn{1}{c}{Center} \\ 
    & & Abs Rel $\downarrow$ & Sq Rel $\downarrow$ & RMSE $\downarrow$ & RMSE$\log$ $\downarrow$ & $\delta < 1.25$ $\uparrow$ & $\delta < 1.25^2$ $\uparrow$ & $\delta < 1.25^3$ $\uparrow$ & OS $\uparrow$ & DS $\uparrow$ & CS $\uparrow$\\ 
    \midrule
    \multirow{3}{*}{GTA} & {1\%} & 0.258 & 4.861 & 10.168 & 0.232 & 0.735 & 0.893 & 0.934 & 0.811 & 0.807 & 0.982 \\ 
    & {10\%} & 0.153 & 3.404 & 6.283 & 0.142 & 0.880 & 0.941 & 0.955 & 0.907 & 0.869 & 0.982 \\
    & {30\%} & 0.134 & 3.783 & 5.165 & 0.117 & 0.908 & 0.948 & 0.957 & 0.932 & 0.891 & 0.982 \\
    & {100\%} & \textbf{0.112} & \textbf{3.361} & \textbf{4.484} & \textbf{0.102} & \textbf{0.923} & \textbf{0.953} & \textbf{0.960} & \textbf{0.953} & \textbf{0.905} & \textbf{0.982} \\ 
    \midrule
    KITTI & {100\%} & 0.074	& 0.449 & 2.847 & 0.126 & 0.954 & 0.980 & 0.987 & 0.962 & 0.918 & 0.974 \\ 
    \bottomrule
    \end{tabular}
}
\vspace{-2mm}
\label{tab:data_amount}
\end{table*}

\begin{figure*}[tpb]
    \small
    \minipage{0.32\textwidth}
        \includegraphics[width=\linewidth, keepaspectratio]{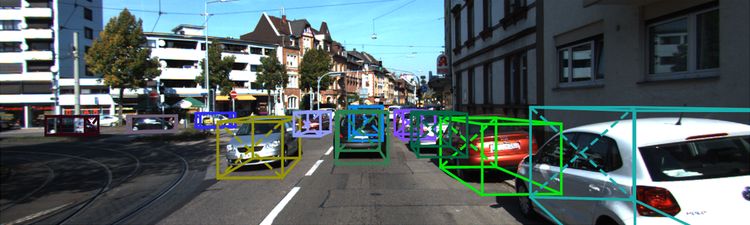}
        \includegraphics[width=\linewidth, keepaspectratio]{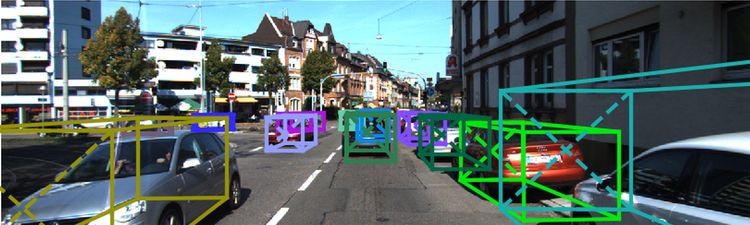}
        \includegraphics[width=\linewidth, keepaspectratio]{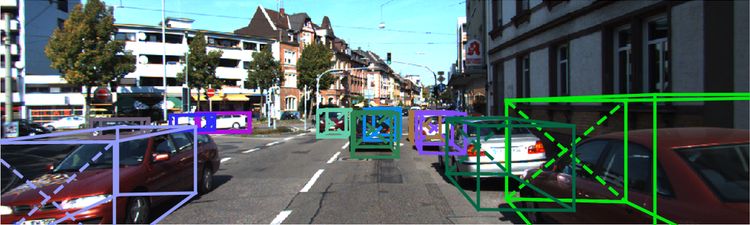}
        \includegraphics[width=\linewidth, keepaspectratio]{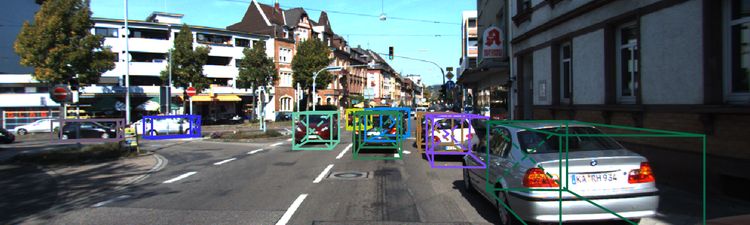}
        \includegraphics[width=\linewidth, keepaspectratio]{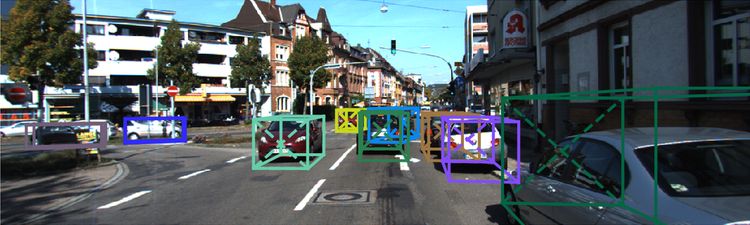}
    \endminipage
    \minipage{0.32\textwidth}
        \includegraphics[width=\linewidth, keepaspectratio]{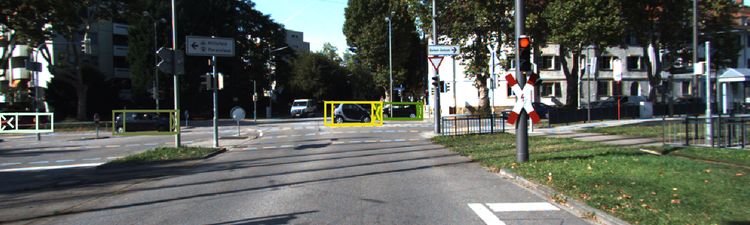}
        \includegraphics[width=\linewidth, keepaspectratio]{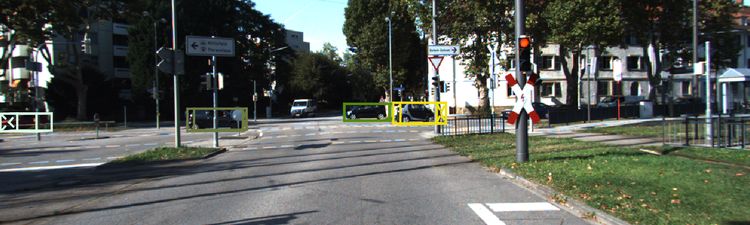}
        \includegraphics[width=\linewidth, keepaspectratio]{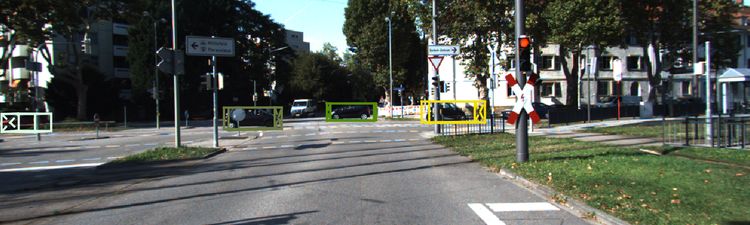}
        \includegraphics[width=\linewidth, keepaspectratio]{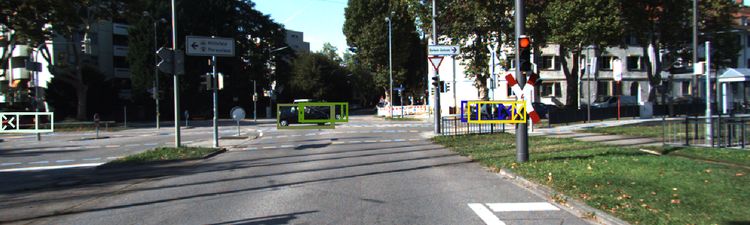}
        \includegraphics[width=\linewidth, keepaspectratio]{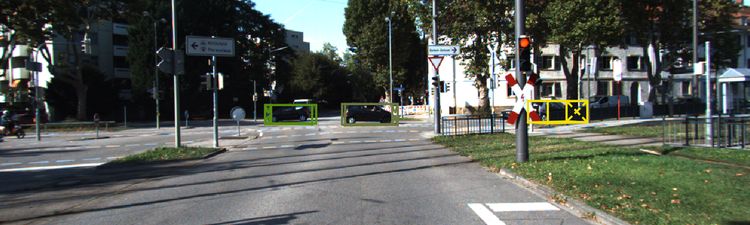}
    \endminipage
    \minipage{0.32\textwidth}
        \includegraphics[width=\linewidth, keepaspectratio]{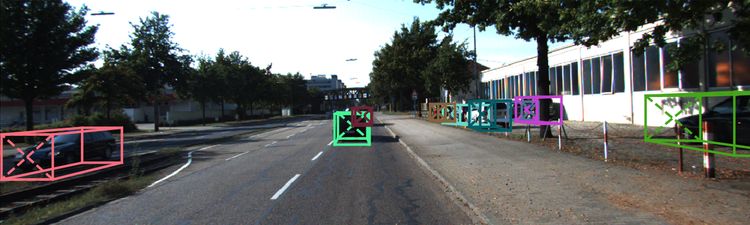}
        \includegraphics[width=\linewidth, keepaspectratio]{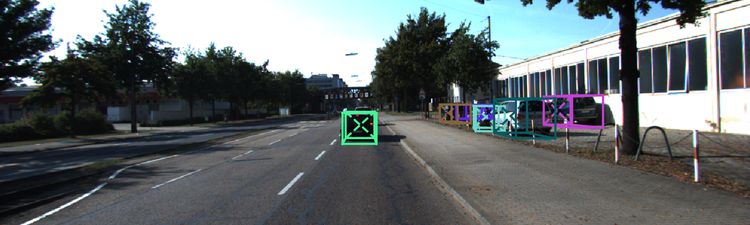}
        \includegraphics[width=\linewidth, keepaspectratio]{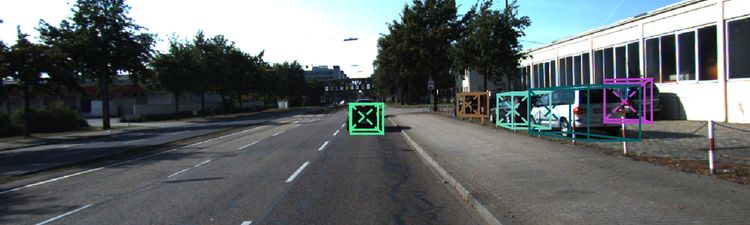}
        \includegraphics[width=\linewidth, keepaspectratio]{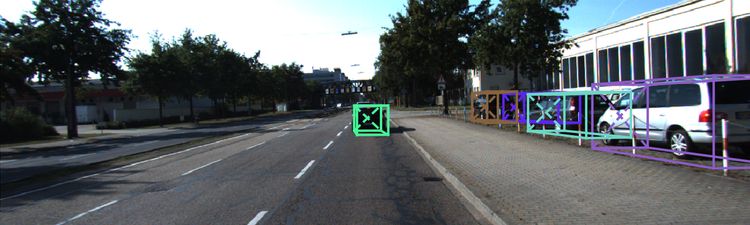}
        \includegraphics[width=\linewidth, keepaspectratio]{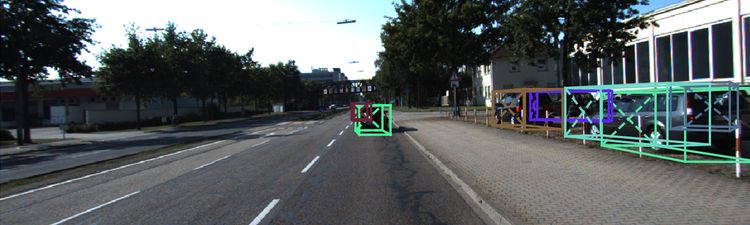}
    \endminipage
    \caption{Qualitative results of 3D Estimation on KITTI testing set.
    We show predicted 3D layout colored with tracking IDs.
    }
    \label{fig:qualitative}
    \vspace{-2mm}
\end{figure*}

\begin{table}[tpb]
\centering
\small
\caption{Tracking performance on the validation set of Argoverse tracking benchmark~\cite{Chang2019argoverse}.
Note that the LiDAR-centric baseline\cite{Chang2019argoverse} uses LiDAR sweeps, HD maps for evaluation.
}
\adjustbox{width=\linewidth}{
    \begin{tabular}{cl|rrrr}
        \toprule
        {Range} & {Method} & {MOTA} $\uparrow$ & {MM} $\downarrow$& {\#FP} $\downarrow$ & {\#FN} $\downarrow$ \\
        \midrule
        \multirow{2}{*}{$30m$} & LiDAR~\cite{Chang2019argoverse} & 73.02 & 19.75 & \textbf{92.80} & \textbf{350.50} \\
         & Ours & \textbf{77.93} & \textbf{5.29} & 104.29 & 395.33 \\
        \midrule
        \multirow{2}{*}{$50m$} & LiDAR~\cite{Chang2019argoverse} & 52.74 & 31.60 & \textbf{99.70} & 1308.25 \\
         & Ours & \textbf{53.48} & \textbf{12.25} & 194.67 & \textbf{857.08} \\
        \midrule
        \multirow{2}{*}{$100m$} &
        LiDAR~\cite{Chang2019argoverse} & \textbf{37.98} & 32.55 & \textbf{105.40} & 2455.30 \\
          & Ours & 15.59 & \textbf{19.83} & 338.54 & \textbf{1572.33} \\
        \bottomrule
    \end{tabular}
}
\label{tab:argo_tracking}
\vspace{-2mm}
\end{table}

\begin{table}[tpb]
\small
\centering
\caption{Tracking performance on the testing set of KITTI tracking benchmark.
Only published methods are reported.
$\dagger$ sign means 3D information is used.
}
\adjustbox{width=\linewidth}{
    \begin{tabular}{lrrrrrr}
        \toprule
        {Methods} & {MOTA} $\uparrow$& {MOTP} $\uparrow$ & {MT} $\uparrow$& {ML} $\downarrow$ & {FP} $\downarrow$ & {FN} $\downarrow$ \\
        \midrule
        Ours & \textbf{84.52} & 85.64 & \textbf{73.38} & \textbf{2.77} & 705 & \textbf{4242} \\
        BeyondPixels$^\dagger$~\cite{MOTBeyondPixels} & 84.24 & \textbf{85.73} & 73.23 & \textbf{2.77} & 705 & 4247 \\
        PMBM$^\dagger$~\cite{Scheidegger2018} & 80.39 & 81.26 & 62.77 & 6.15 & 1007 & 5616 \\
        MDP~\cite{xiang2015mdptrack} & 76.59 & 82.10 & 52.15 & 13.38 & \textbf{606} & 7315 \\
        \bottomrule
    \end{tabular}
}
\label{tab:kitti_tracking}
\vspace{-2mm}
\end{table}

\minisection{3D for tracking.}
The ablation study of tracking performance could be found in \tabref{tab:gta_tracking}.
Adding deep feature distinguishes two near-overlapping objects, our false negative (FN) rate drops with an observable margin. 
With depth-order matching and occlusion-aware association, our model filters out $6-8\%$ possible mismatching trajectories. 
For a full ablation study, please refer to the \apnref{apn:experiments}


\minisection{Motion Modeling.}
We propose to use LSTM to model the vehicle motion. To analyze its effectiveness, we compare our LSTM model with traditional 3D Kalman filter (KF3D) and single frame 3D estimation using 3D IoU mAP.
\tabref{tab:tracking_iou} shows that KF3D provides a small improvement via trajectory smoothing within prediction and observation.
On the other hand, our LSTM module gives a learned estimation based on past $n$ velocity predictions and current frame observation, which may compensate for the observation error.
Our LSTM module achieves the highest accuracy among the other methods with all the IoU thresholds.

\minisection{3D Center Projection Estimation.} We estimate the 3D location of a bounding box through predicting the projection of its center and depth, while Mousavian~\etal~\cite{mousavian20173d} uses the center of detected 2D boxes directly. \tabref{tab:projection_3d_center} shows the comparison of these two methods on KITTI dataset. The result indicates the correct 3D projections provides higher tracking capacity for motion module to associate candidates and reduces the ID switch (IDS) significantly.

\minisection{Amount of Data Matters.}
We train the depth estimation module with $1\%$, $10\%$, $30\%$ and $100\%$ training data.
The results show how we can benefit from more data in \tabref{tab:data_amount}, where there is a consistent trend of performance improvement as the amount of data increases.
The trend of our results with a different amount of training data indicates that large-scale 3D annotation is helpful, especially with accurate ground truth of far and small objects.

\minisection{Real-world Evaluation.}
Besides evaluating on synthetic data, we resort to Argoverse~\cite{Chang2019argoverse} and KITTI~\cite{geiger2012we} tracking benchmarks to compare our model abilities.
For Argoverse, we use Faster R-CNN detection results of mmdetection~\cite{mmdetection} implementation pre-trained on COCO~\cite{lin2014mscoco} dataset.
Major results are listed in \tabref{tab:argo_tracking} and \tabref{tab:kitti_tracking}.
For a full evaluation explanation, please refer to the \apnref{apn:experiments}.
Our monocular 3D tracking method outperforms all the published methods on KITTI and beats LiDAR-centric baseline methods on $50m$ and $30m$ ranges of the Argoverse tracking validation set upon the time of submission.

It is worth noting that the baseline methods on Argoverse tracking benchmark leveraging HD maps, locations, and LiDAR points for 3D detection, in addition to images. Our monocular 3D tracking approach can reach competitive results with image stream only. It is interesting to see that the 3D tracking accuracy based on images drops much faster with increasing range threshold than LiDAR-based method. This is probably due to different error characteristics of the two measurement types. The farther objects occupy smaller number of pixels, leading to bigger measurement variance on the images. However, the distance measurement errors of LiDAR hardly change for faraway objects. At the same time, the image-based method can estimate the 3D positions and associations of nearby vehicles accurately. The comparison reveals the potential research directions to combine imaging and LiDAR signals.
 


%% file: 06Conclusion.tex
\section{Conclusion}
In this paper, we learn 3D vehicle dynamics from monocular videos.
We propose a novel framework, combining spatial visual feature learning and global 3D state estimation, to track moving vehicles in a 3D world.
Our pipeline consists of a single-frame monocular 3D object inference model and motion LSTM for inter-frame object association and updating.
In data association, we introduced occlusion-aware association to solve inter-object occlusion problem.
In tracklet matching, depth ordering filters out distant candidates from a target.
The LSTM motion estimator updates the velocity of each object independent of camera movement.
Both qualitative and quantitative results indicate that our model takes advantage of 3D estimation leveraging our collection of dynamic 3D trajectories.

%% file: 07Acknowledgements.tex
\section{Acknowledgements}
The authors gratefully acknowledge the support of Berkeley AI Research, Berkeley DeepDrive and MOST-107 2634-F-007-007, MOST Joint Research Center for AI Technology and All Vista Healthcare.

%% file: appendix.tex
\appendix

\section*{Foreword}
This appendix provides technical details about our monocular 3D detection and tracking network and our virtual GTA dataset, and more qualitative and quantitative results.
\secref{apn:3d_center_projection} shows the importance of learning a 3D center projection. 
\secref{apn:data_association} provides more details about occlusion aware association.
\secref{apn:motion_model} will talk about details of two LSTMs predicting and updating in 3D coordinates.
\secref{apn:data_stats} offers frame- and object-based statistical summaries.
\secref{apn:training_detail} describes our training procedure and network setting of each dataset.
\secref{apn:experiments} illustrates various comparisons, including ablation, inference time and network settings, of our method on GTA, Argoverse and KITTI data.

\section{Projection of 3D Center}
\label{apn:3d_center_projection}
\begin{figure}[htpb]
	\begin{center}
		\includegraphics[width=1.0\linewidth]{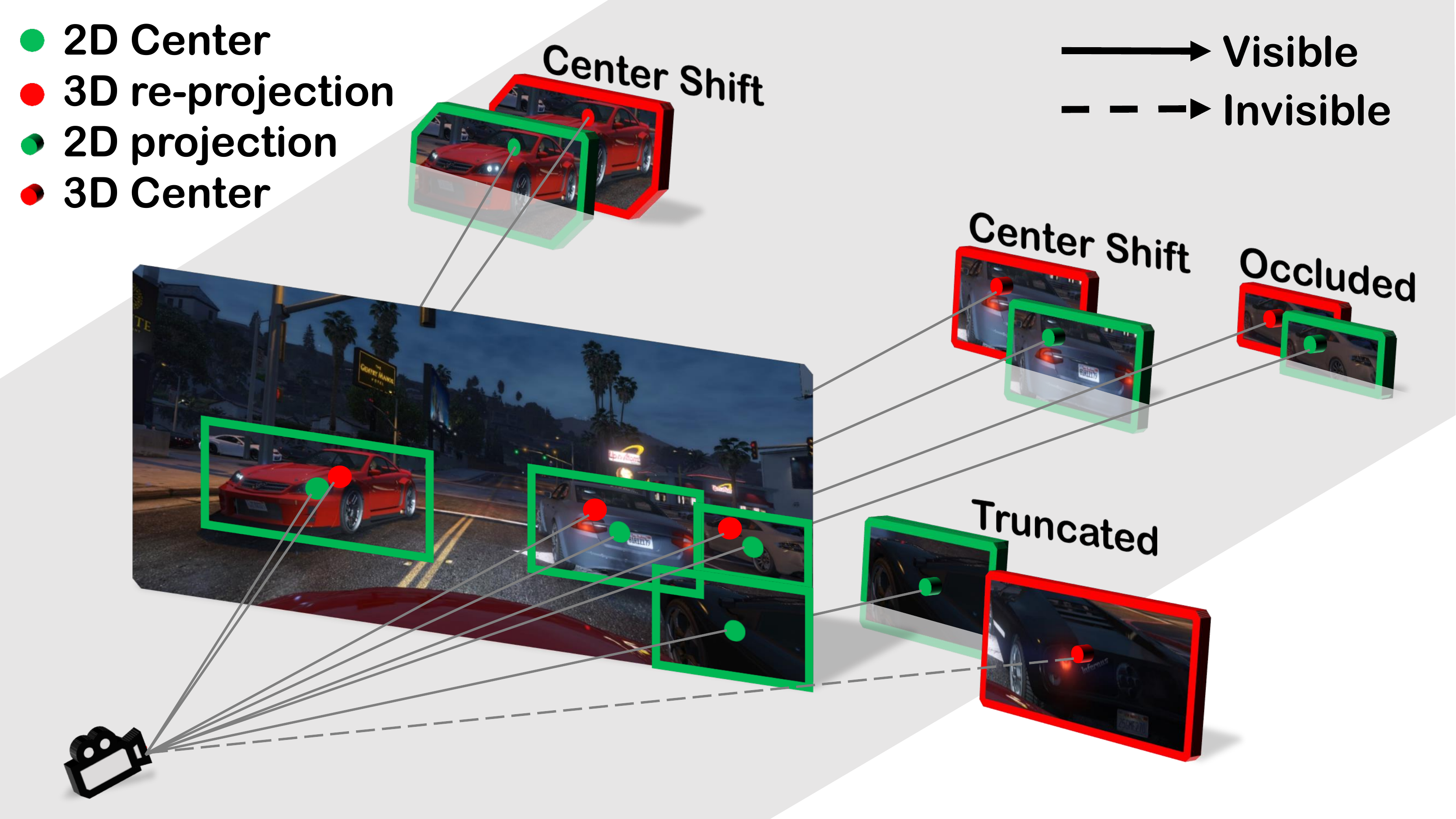}
	\end{center}
	\caption{The illustration of the difference in 2D bounding box and 3D projected centers. 
	A visible 2D center projection point may wrongly locate above or under the ground plane in the 3D coordinates. 
	Other states, occluded or truncated, would inevitably suffer from the center shift problem.
	The center shift causes misalignment for GT and predicted tracks and harms 3D IOU AP performance.}
	\label{fig:3d_center}
\end{figure}

We estimate the vehicle 3D location by first estimating the depth of the detected 2D bounding boxes. Then the 3D position is located based on the observer's pose and camera calibration. We find that accurately estimating the vehicle 3D center and its projection is critical for accurate 3D box localization. However, the 2D bounding box center can be far away from the projection of 3D center for several reasons. First, there is always a shift from the 2D box center if the 3D bounding box is not axis aligned in the observer's local coordinate system. Second, the 2D bounding box is only detecting the visible area of the objects after occlusion and truncation, while the 3D bounding box is defined on the object's full physical dimensions. The projected 3D center can be even out of the detected 2D boxes. For instance, the 3D bounding box of a truncated object can be out of the camera view. These situations are illustrated in \figref{fig:3d_center}. 

\section{Data Association Details}
\label{apn:data_association}

Due to the length limitation in the main paper, we supplement more details of data association here.

\minisection{Occlusion-Aware Association.}
We continue to predict the 3D location of unmatched tracklets until they disappear from our tracking range (\eg $0.15m$ to $100m$) or die out after $20$ time-steps.
The sweeping scheme benefits tracker from consuming a huge amount of memory to store all the $\mathtt{occluded}$ tracks.

\minisection{Depth-Ordering Matching.}
The equation mentioned in the main paper
\begin{equation}
\begin{split}
    \mathbf{A}_{\subfix{3D}}(\mathbf{\tau}_a, \mathbf{s}_a) &= \mathds{1}\times\frac{\phi(\mf{M}(X_{\mathbf{\tau}_a})) \cap \mf{M}(X_{\mathbf{s}_a})}
    {\phi(\mf{M}(X_{\mathbf{\tau}_a})) \cup \mf{M}(X_{\mathbf{s}_a})},
\end{split}
\end{equation}
where $$\mathds{1}=(d_{\mathbf{\tau}_a}-d_{\mathbf{\mathbf{s}_a}}) < l_{\mathbf{\tau}_a} + w_{\mathbf{\tau}_a} + l_{\mathbf{s}_a} + w_{\mathbf{s}_a}$$ denotes if the tracklets is kept after depth filtering, which can be interpreted as a loose bound 
$$ \sqrt{l_{\mathbf{\tau}_a}^2 + w_{\mathbf{\tau}_a}^2} + \sqrt{l_{\mathbf{s}_a}^2 + w_{\mathbf{s}_a}^2} < l_{\mathbf{\tau}_a} + w_{\mathbf{\tau}_a} + l_{\mathbf{s}_a} + w_{\mathbf{s}_a}$$ of two car within a reachable distance range, \ie the diagonal length of two car.
The overlapping function 
$$\phi(\cdot)=\argmin_x \{x | \mathtt{ord}(x) < \mathtt{ord}(x_0) \forall x_0 \in \mathtt{M}(X_{\mathbf{\tau}_a}))\} $$ captures pixels of non-occluded tracklets region with the nearest depth order $\mathtt{ord(\cdot)}$.
Note that the $\mathtt{ord(\cdot)}$ returns the same order if the distance is within $1$ meter.
Tracklets closer to the DOI occlude the overlapped area of farther ones.
So each DOI matches to the tracklet with highest IOU while implicitly modeling a strong prior of depth sensation called ``Occultation''.

\section{Motion Model}
\label{apn:motion_model}

\minisection{Kalman Filter.}
In our experiments, we compare a variant of approaches in estimating locations.
2D Kalman filter (KF2D) models $$\{x, y, s, a, \Delta x, \Delta y, \Delta a\}$$ in 2D coordinates, where $s$ stands for the width/high ratio and $a$ for the area of the object.
KF2D often misses tracking due to substantial dynamic scene changes from sudden ego-motion.
We also propose a 3D Kalman filter (KF3D) which models $$\{x, y, z, \Delta x, \Delta y, \Delta z\}$$ in 3D coordinates.
However, Kalman filter tries to find a position in between observation and prediction location that minimize a mean of square error, while LSTM does not be bounded in such a restriction.

\minisection{Deep Motion.}
The proposed final motion model consists of two LSTMs, one for predicting and the other for updating.
The latter is in charge of refining 3D locations.

Predicting LSTM (P-LSTM) learns a velocity in 3D $\{\Delta x, \Delta y, \Delta z\}$ using hidden state based on the previously $n$ updated velocities $\dot{P}_{T-n:T-1}$ and location $\overline{P}_{T-1}$ and predicts a smoothed step toward an estimated location $\tilde{P}_T-1$.
We embed velocity for the past $n=5$ consecutive frames with an initial value of all $0$ to model object motion and acceleration from the trajectory.
As the next time step comes, the velocity array updates a new velocity into the array, and discards the oldest one.

For Updating LSTM (U-LSTM) module, we encode both predicted location $\tilde{P}_{T-1}$ and current single-frame observation $\hat{P}_{T}$ into two location features.
By concatenating them together, we obtained a $128$-dim feature as the input of U-LSTM.
The LSTM learns a smooth estimation of location difference from the observation $\hat{P}_{T}$ to the predicted location $\tilde{P}_{T-1}$ and updates the 3D location  $\overline{P}_T$ and velocity $\dot{P}_{T-n+1:T}$ from hidden state leveraging previous observations.

Both LSTM modules are trained with ground truth location projected using the depth value, a projection of 3D center, camera intrinsic and extrinsic using $2$ losses.
The L1 loss $\mathtt{L}_1=|\dot{P}_T - \dot{P}_{T-1}|$ reduces the distance of estimated and ground truth location.
The linear motion loss $\mathtt{L}_{linear}=(1-cos(\dot{P}_T, \dot{P}_{T-1})) + L1(\dot{P}_T, \dot{P}_{T-1})$ focuses on the smooth transition of location estimation.

\begin{table*}[htpb]
    \caption{Comparison to related dataset for detection and tracking (Upper half: real-world, Lower half: synthetic). We only count the size and types of annotations for training and validation (D=detection, T=tracking, C=car, P=pedestrian). To our knowledge, our dataset is the largest 3D tracking benchmark for dynamic scene understanding, with control signals of driving, sub-categories of object.} 
	\label{tab:dataset}
	\centering
\adjustbox{width=0.8\linewidth}{
	\begin{tabular}{cccccccccc}
	\toprule
		Dataset & Task & Object & Frame & Track & Boxes & 3D & Weather & Occlusion & Ego-Motion\\ \midrule
		KITTI-D~\cite{geiger2012we} & D & C,P & 7k & - & 41k & \checkmark & - & \checkmark & - \\
		KAIST~\cite{hwang2013multispectral} & D & P & 50k & - & 42k & - & \checkmark & \checkmark & - \\
		KITTI-T~\cite{geiger2012we} & T & C & 8k & 938 & 65k & \checkmark & - & \checkmark & \checkmark \\
		MOT15-3D~\cite{leal2015motchallenge} & T & P & 974 & 29 & 5632 & \checkmark & \checkmark & - & - \\
		MOT15-2D~\cite{leal2015motchallenge} & T & P & 6k & 500 & 40k & - & \checkmark & - & - \\
		MOT16~\cite{milan2016mot16} & T & C,P & 5k & 467 & 110k & - & \checkmark & \checkmark & - \\
		UA-DETRAC~\cite{wen2015ua} & D,T & C & 84k & 6k & 578k & - & \checkmark & \checkmark & - \\
		\midrule
		Virtual KITTI~\cite{gaidon2016virtual} & D,T & C & 21k & 2610 & 181k & \checkmark & \checkmark & \checkmark & \checkmark \\
		VIPER~\cite{richter2017playing} & D,T & C,P & 184k & 8272 & 5203k & \checkmark & \checkmark & \checkmark & \checkmark  \\
		Ours & D,T & C,P & 688k & 325k & 10068k & \checkmark & \checkmark & \checkmark & \checkmark \\ 
		\bottomrule
	\end{tabular} 
	}
\end{table*}

\section{Dataset Statistics}
\label{apn:data_stats}
To help understand our dataset and its difference, we show more statistics.

\minisection{Dataset Comparison.}
\tabref{tab:dataset} demonstrates a detailed comparison with related datasets, including detection, tracking, and driving benchmarks. 
KITTI-D~\cite{geiger2012we} and KAIST~\cite{hwang2013multispectral} are mainly detection datasets, while KITTI-T~\cite{geiger2012we}, MOT15~\cite{leal2015motchallenge}, MOT16~\cite{milan2016mot16}, and UA-DETRAC~\cite{wen2015ua} are primarily 2D tracking benchmarks. 
The common drawback could be the limited scale, which cannot meet the growing demand for training data. 
Compared to related synthetic dataset, Virtual KITTI~\cite{gaidon2016virtual} and VIPER~\cite{richter2017playing}, we additionally provide fine-grained attributes of object instances, such as color, model, maker attributes of vehicle, motion and control signals, which leaves the space for imitation learning system. 

\begin{figure}[htpb]
    \minipage{0.49\linewidth}
        \includegraphics[width=1.0\linewidth, keepaspectratio]{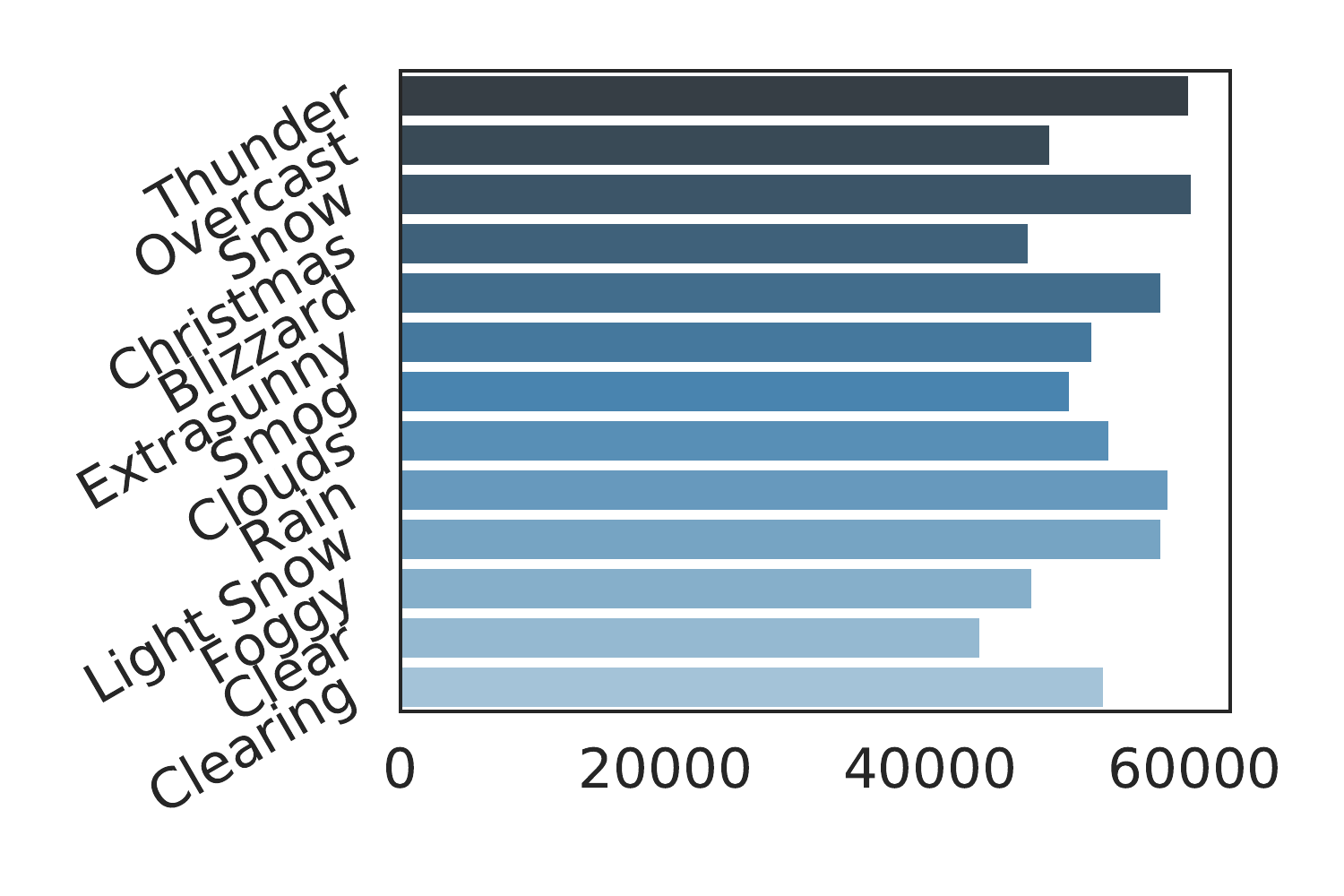}
        \subcaption{Weather}\label{fig:statistics_weather}
    \endminipage
    \minipage{0.49\linewidth}
        \includegraphics[width=1.0\linewidth, keepaspectratio]{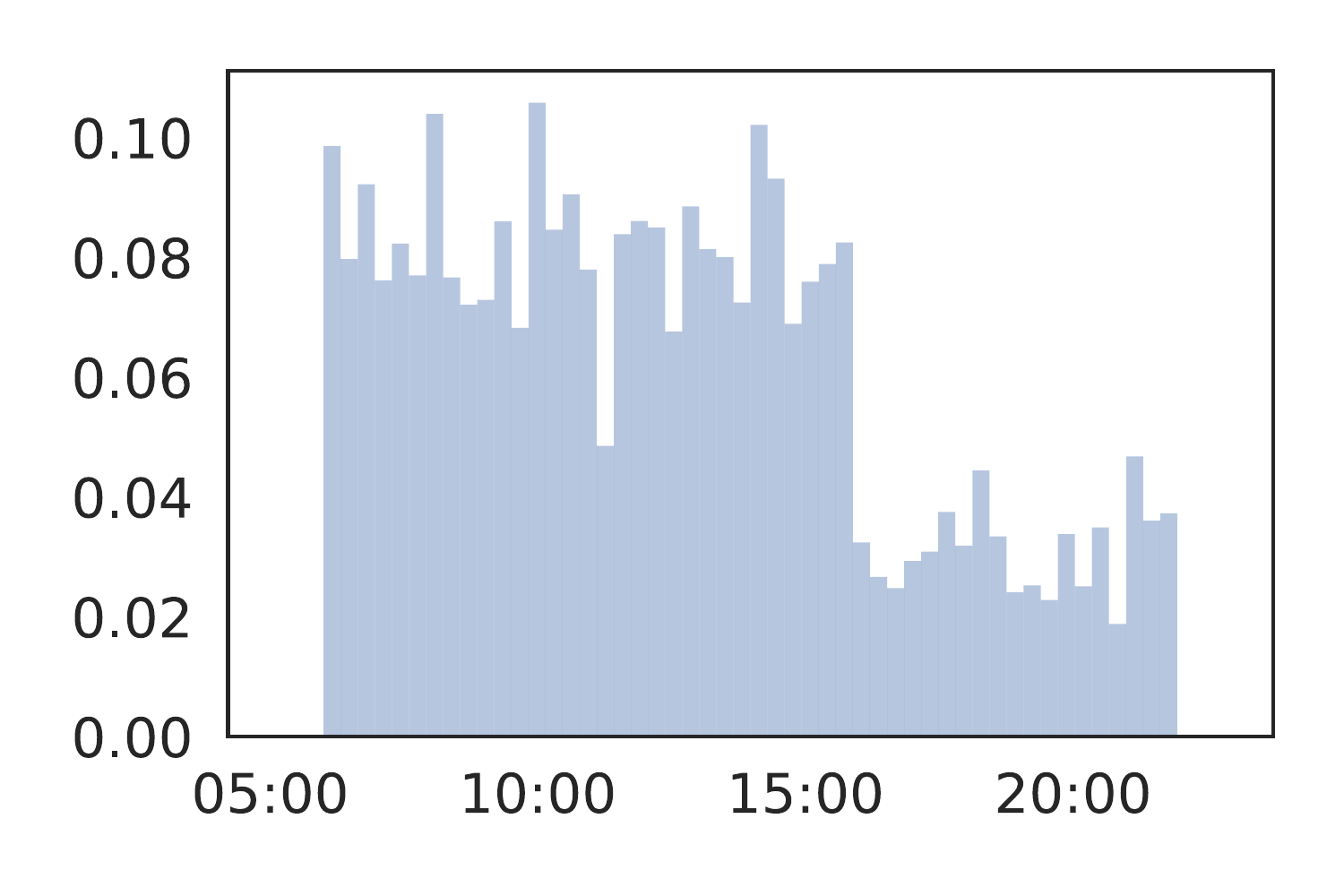}
        \subcaption{Hours}\label{fig:statistics_hours}
    \endminipage
    \caption{The statistics of scene in our dataset.}
    \label{fig:statistics_scene}
\end{figure}

\minisection{Weather and Time of Day.}
\figref{fig:statistics_scene} shows the distribution of weather, hours of our dataset.
It features a full weather cycle and time of a day in a diverse virtual environment.
By collecting various weather cycles (\figref{fig:statistics_weather}), our model learns to track with a higher understanding of environments.
With different times of a day (\figref{fig:statistics_hours}), the network handles changeable perceptual variation.

\begin{figure}[htpb]
    \minipage{1.0\linewidth}
        \includegraphics[width=1.0\linewidth, keepaspectratio]{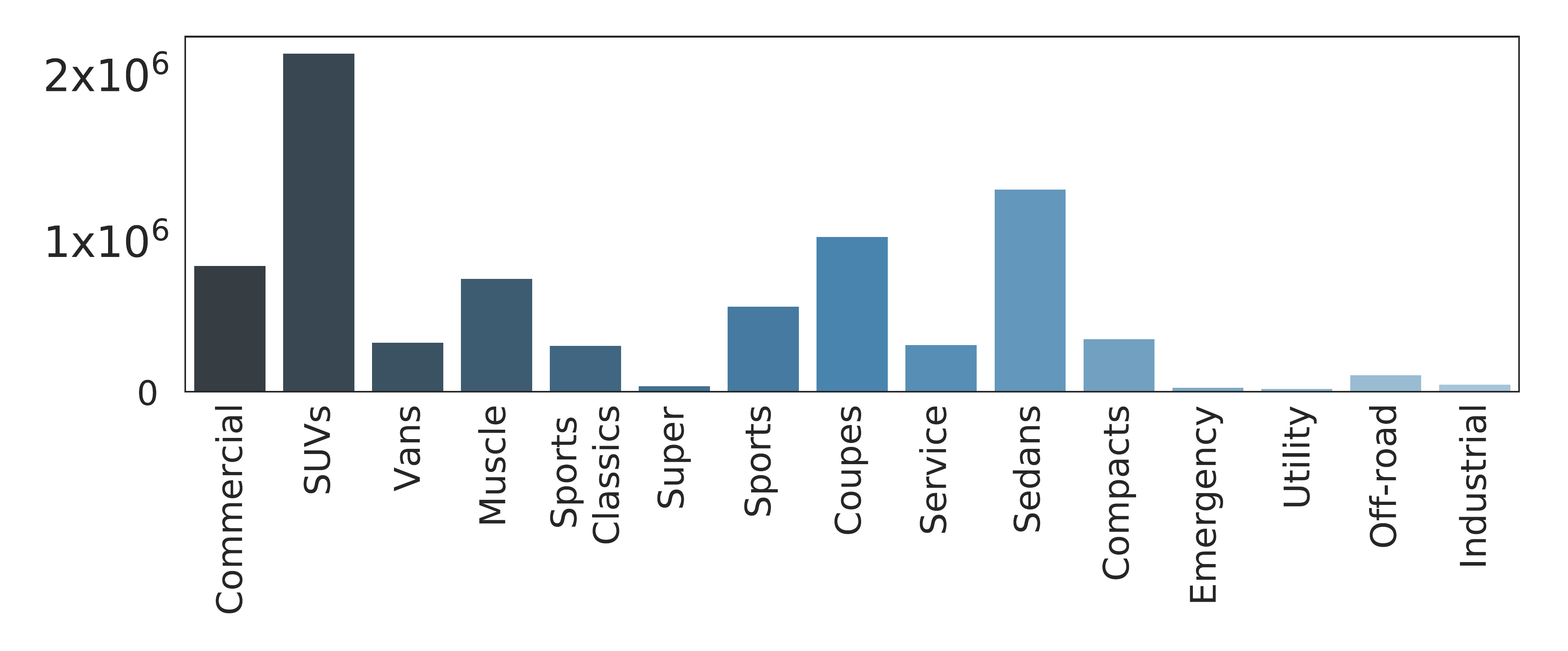}
        \subcaption{Category}\label{fig:statistics_category}
    \endminipage
    \hfill
    \minipage{0.49\linewidth}
        \includegraphics[width=1.0\linewidth, keepaspectratio]{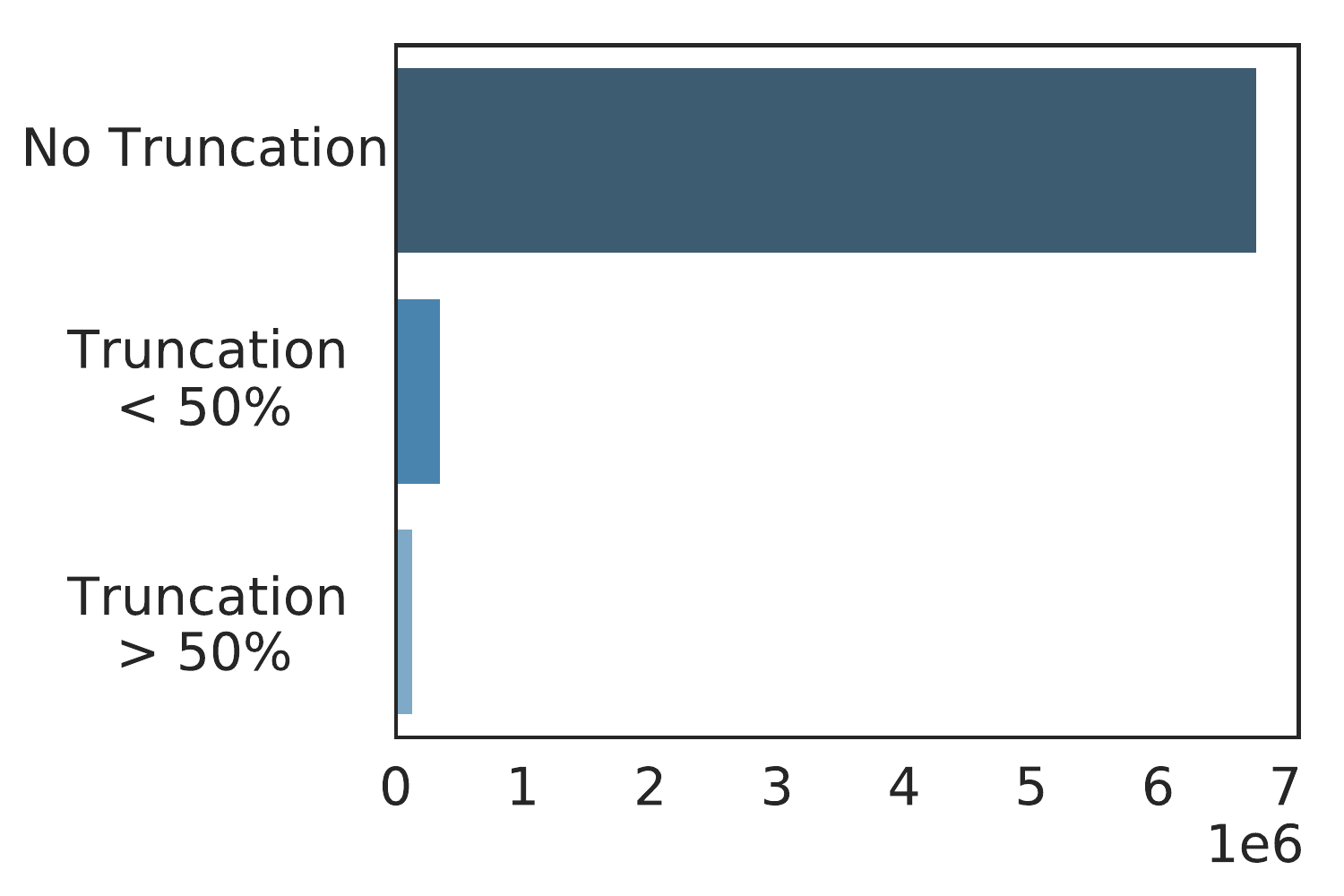}
        \subcaption{Truncation}\label{fig:statistics_truncated}
    \endminipage
    \minipage{0.49\linewidth}
        \includegraphics[width=1.0\linewidth, keepaspectratio]{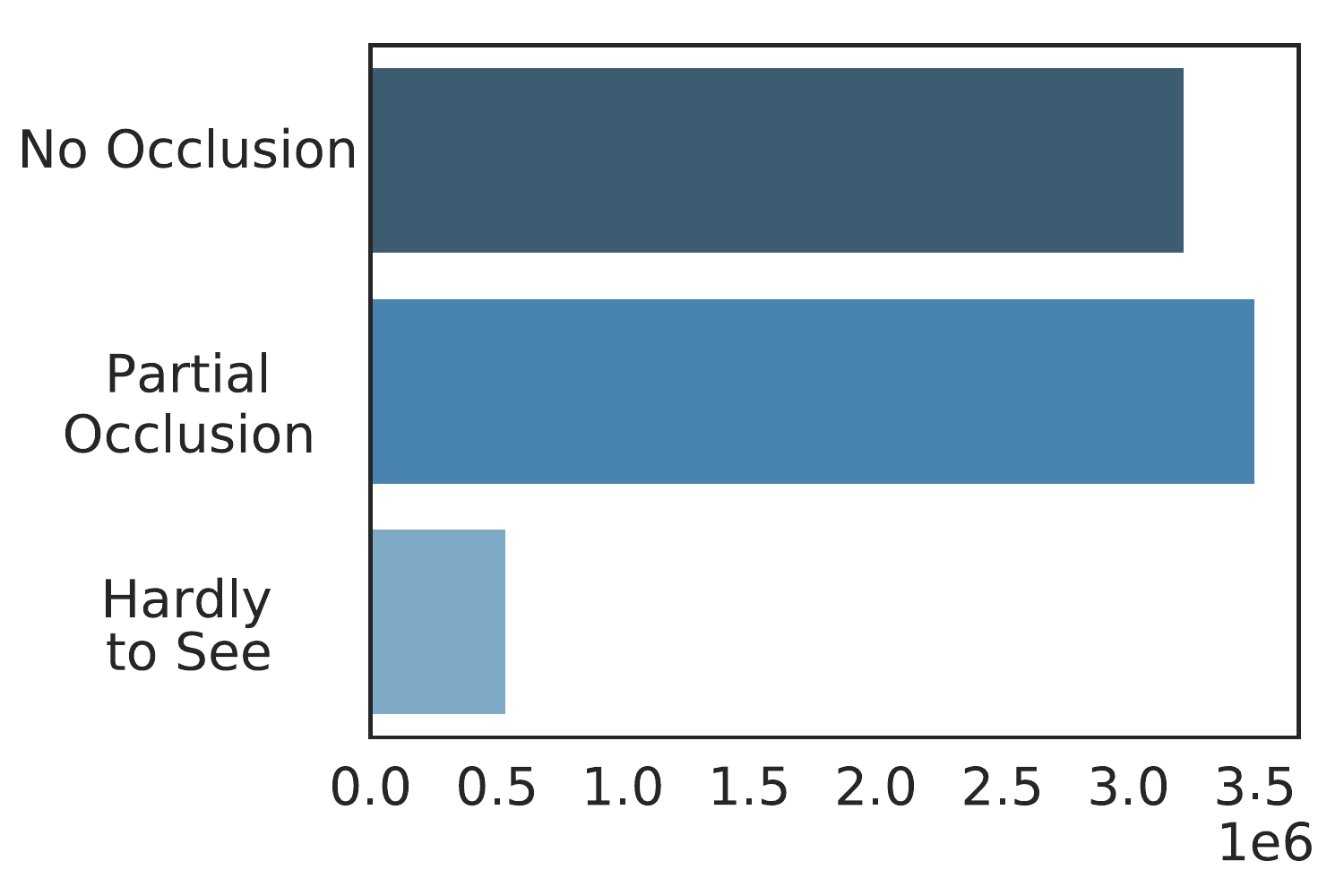}
        \subcaption{Occlusion}\label{fig:statistics_occlusion}
    \endminipage
	\caption{The statistics of object in our dataset.}
	\label{fig:statistics_object}
\end{figure}

\minisection{Number of Instances in Each Category.}
The vehicle diversity is also very large in the GTA world, featuring $15$ fine-grained subcategories. We analyzed the distribution of the $15$ subcategories in \figref{fig:statistics_category}.
Besides instance categories, we also show the distribution of occluded (\figref{fig:statistics_occlusion}) and truncated (\figref{fig:statistics_truncated}) instances to support the problem of partially visible in the 3D coordinates.

\begin{figure}[htpb]
    \begin{subfigure}{0.45\linewidth}
		\includegraphics[width=\linewidth]{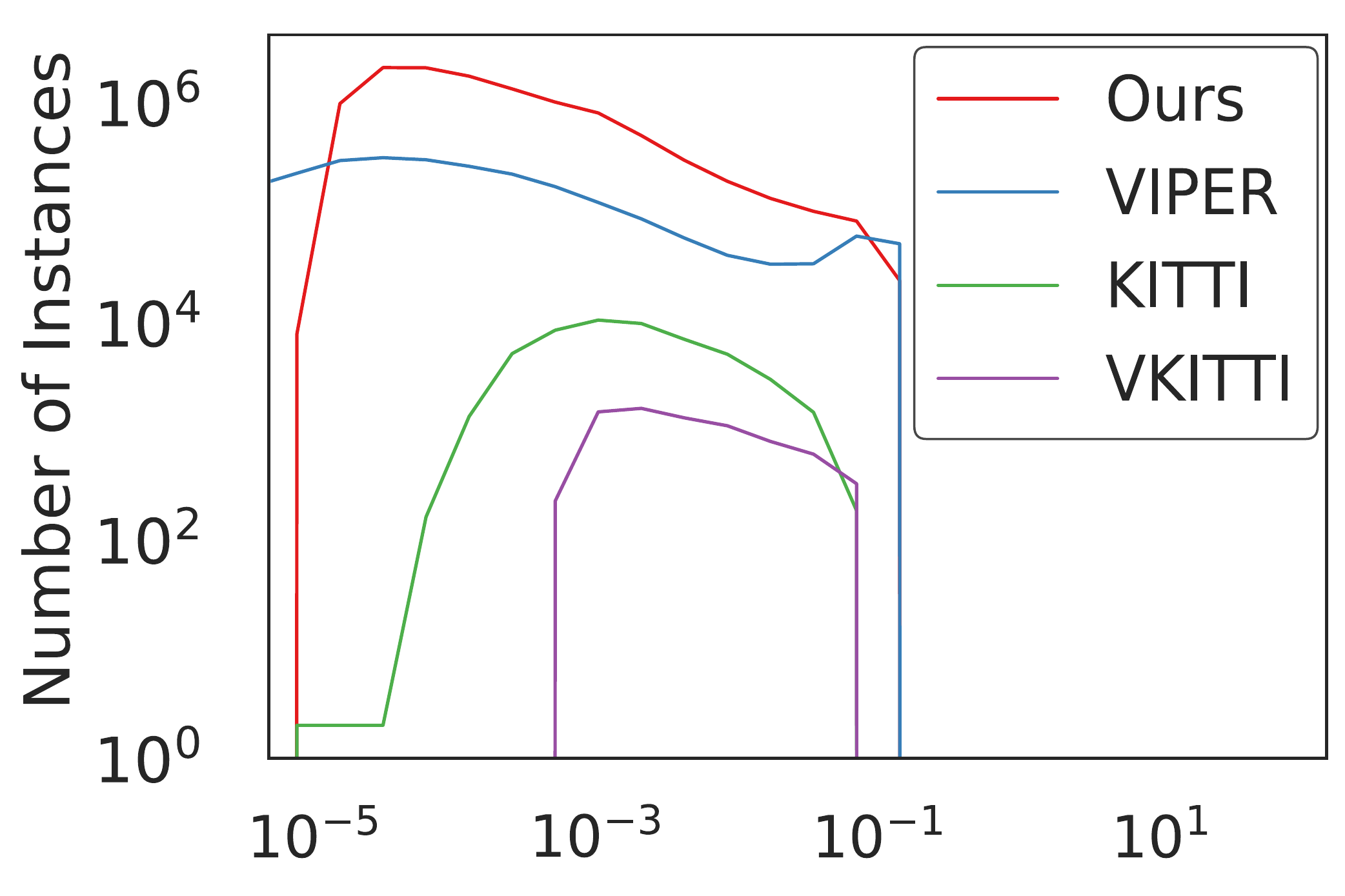}
		\caption{Instance scale}\label{fig:statistics_scale}
	\end{subfigure}
	\begin{subfigure}{0.45\linewidth}
		\includegraphics[width=\linewidth]{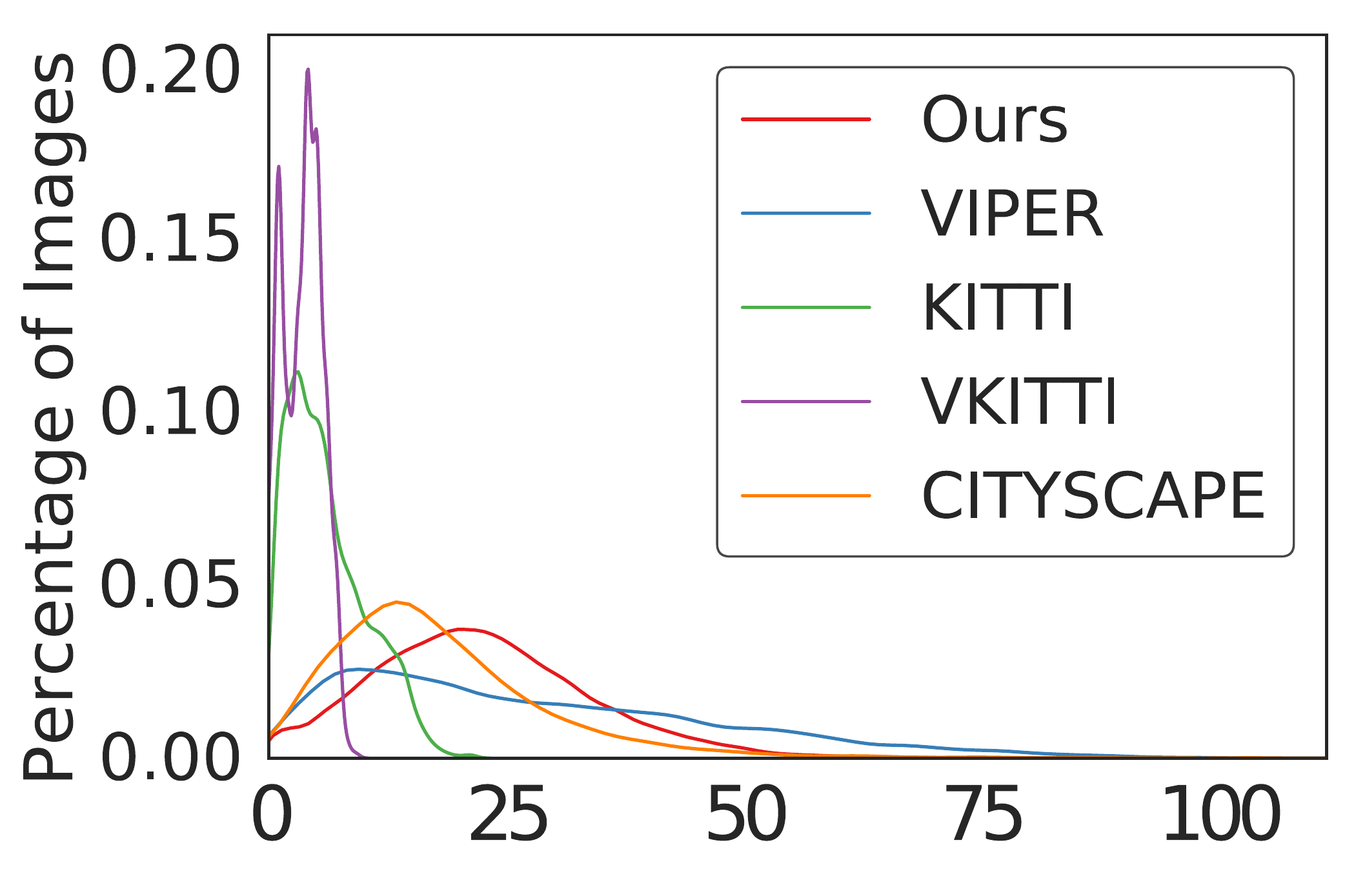}
		\caption{Instances per image}\label{fig:statistics_ins}
	\end{subfigure}
    \begin{subfigure}{0.45\linewidth}
		\includegraphics[width=\linewidth]{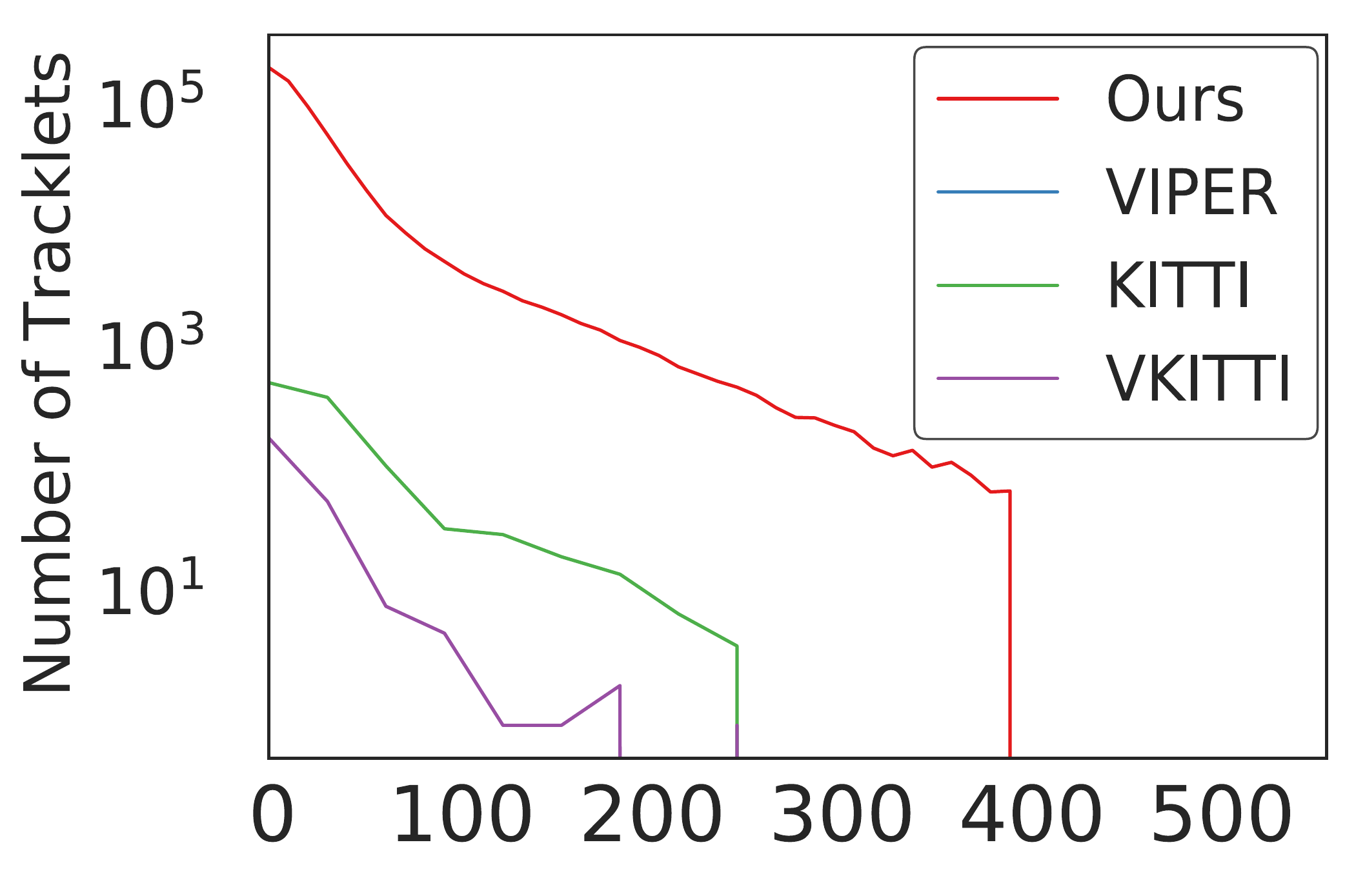}
		\caption{Frames of tracking}\label{fig:statistics_duration}
	\end{subfigure}
	\begin{subfigure}{0.45\linewidth}
		\includegraphics[width=\linewidth]{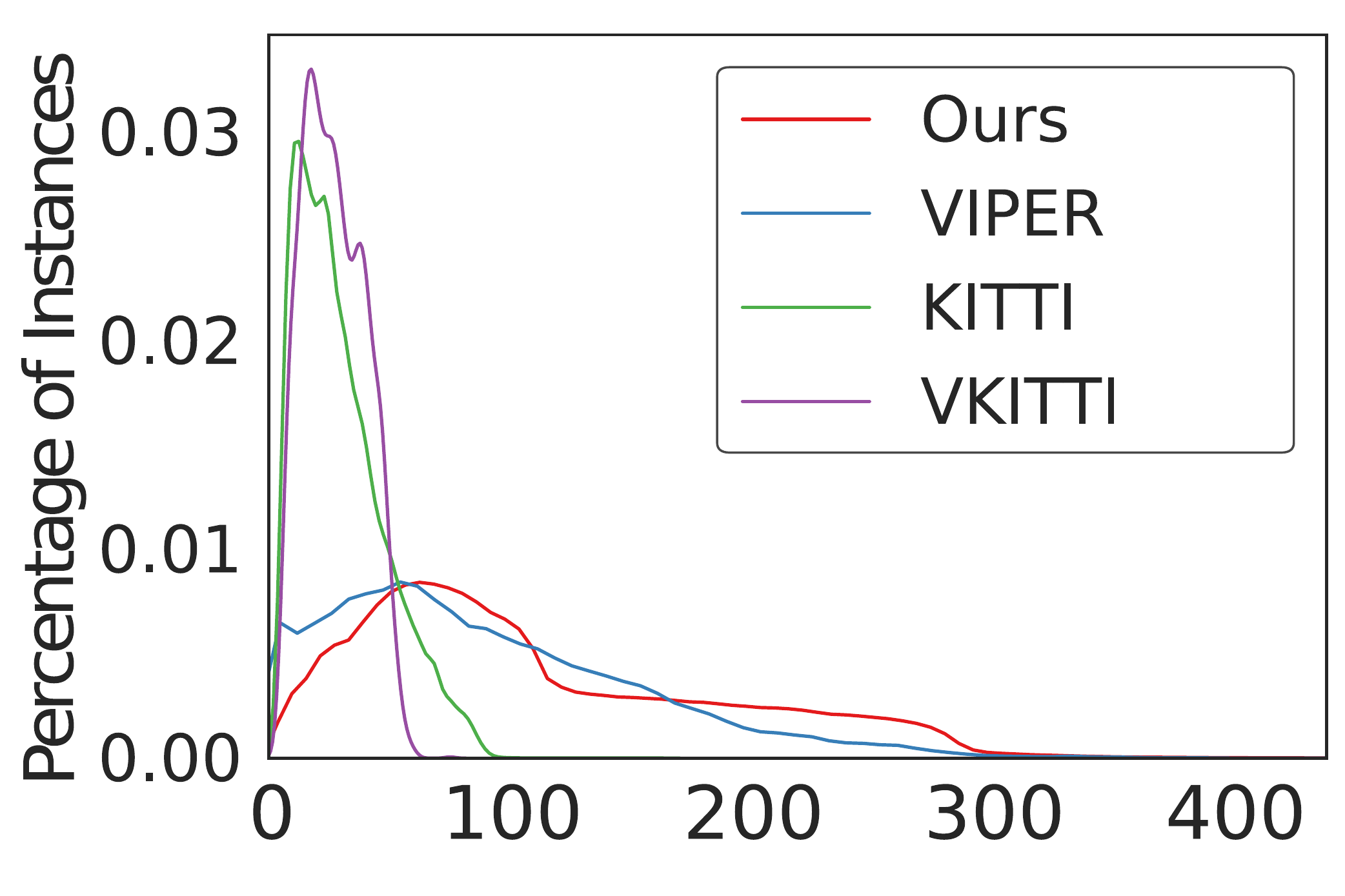}
		\caption{Vehicle distance}\label{fig:statistics_distance}
	\end{subfigure}
    \caption{Statistical summary of our dataset in comparison of KITTI~\cite{geiger2012we}, VKITTI~\cite{gaidon2016virtual}, VIPER~\cite{richter2017playing}, and Cityscapes\cite{cordts2016cityscapes}}
    \label{fig:statistics}
\end{figure}

\minisection{Dataset Statistics.}
Compared to the others, our dataset has more diversity regarding instance scales (\figref{fig:statistics_scale}) and closer instance distribution to real scenes (\figref{fig:statistics_ins}).

\begin{figure*}[htpb]
    \adjustbox{width=1.0\linewidth}{
    \setlength{\tabcolsep}{2pt}
    \begin{tabular}{rl}
        \includegraphics[width=0.49\linewidth, keepaspectratio]{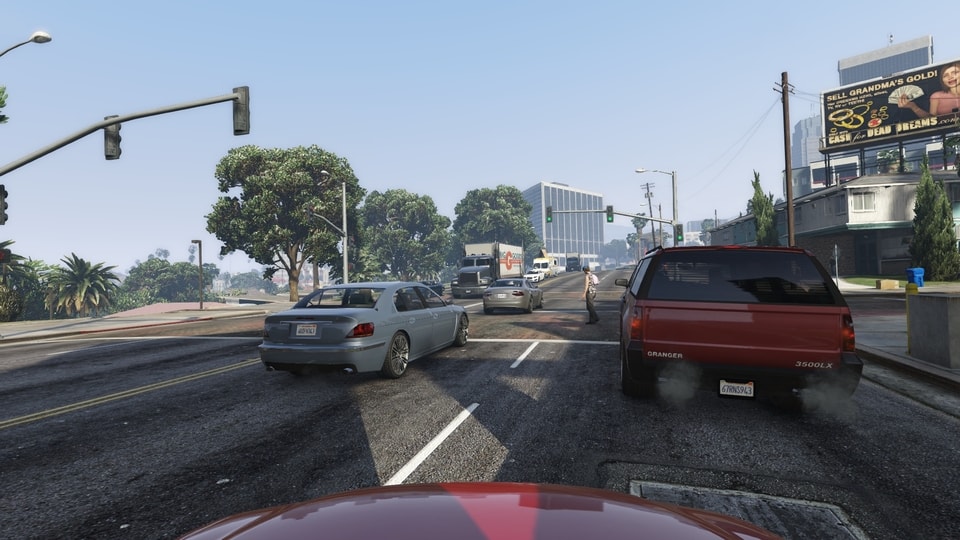} &
        \includegraphics[width=0.49\linewidth, keepaspectratio]{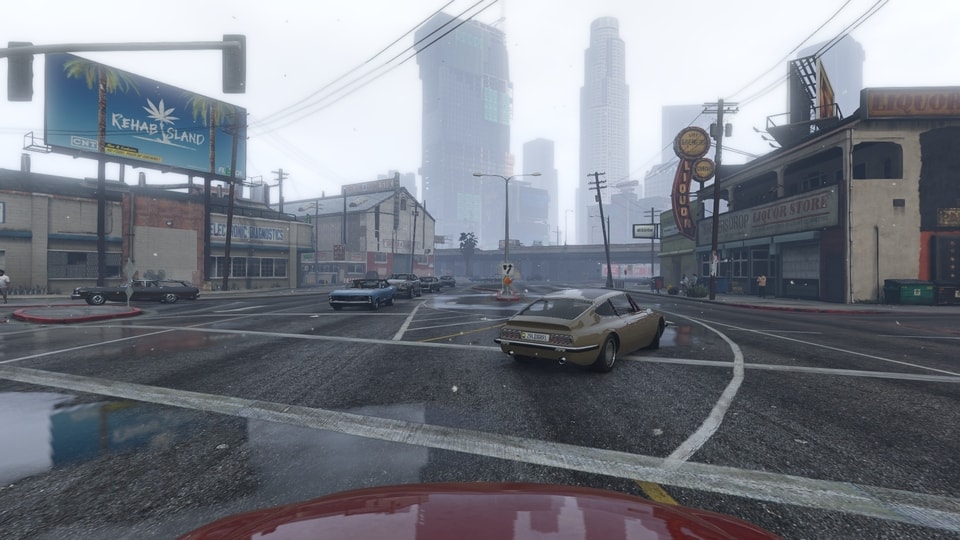} \\
        \includegraphics[width=0.49\linewidth, keepaspectratio]{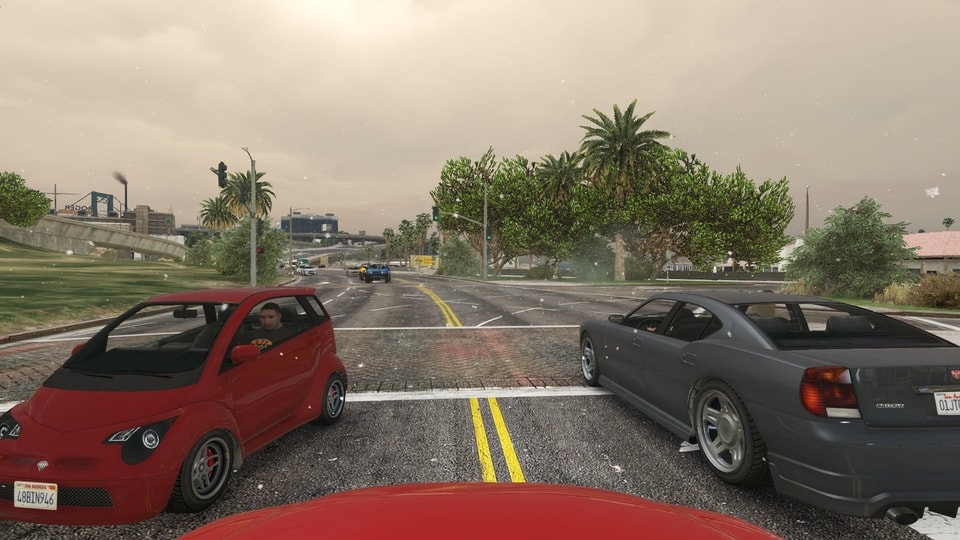} &
        \includegraphics[width=0.49\linewidth, keepaspectratio]{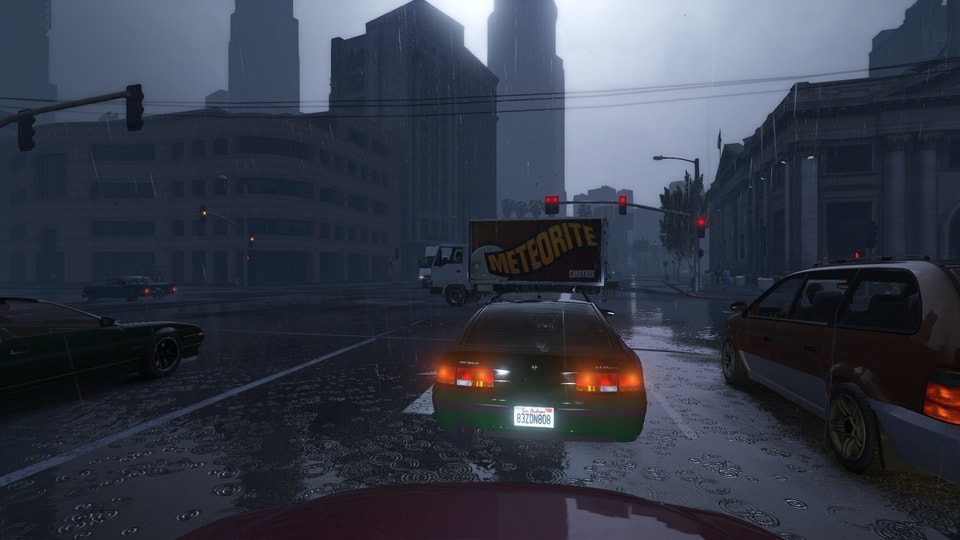} \\
        \includegraphics[width=0.49\linewidth, keepaspectratio]{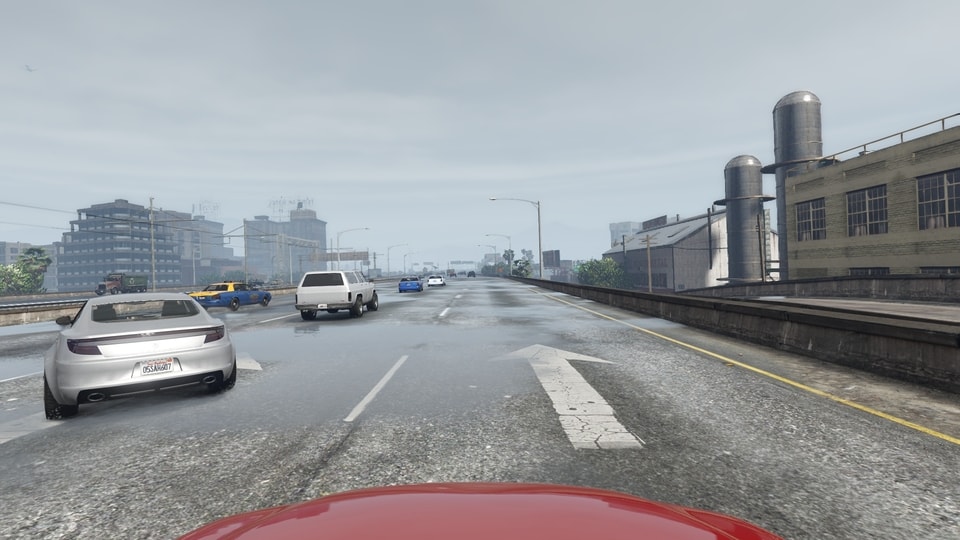} &
        \includegraphics[width=0.49\linewidth, keepaspectratio]{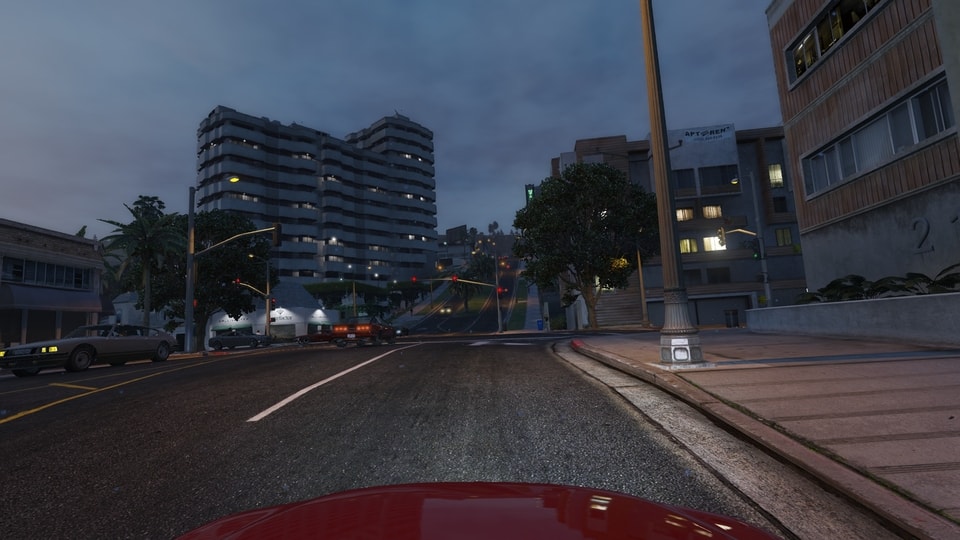} \\
        \includegraphics[width=0.49\linewidth, keepaspectratio]{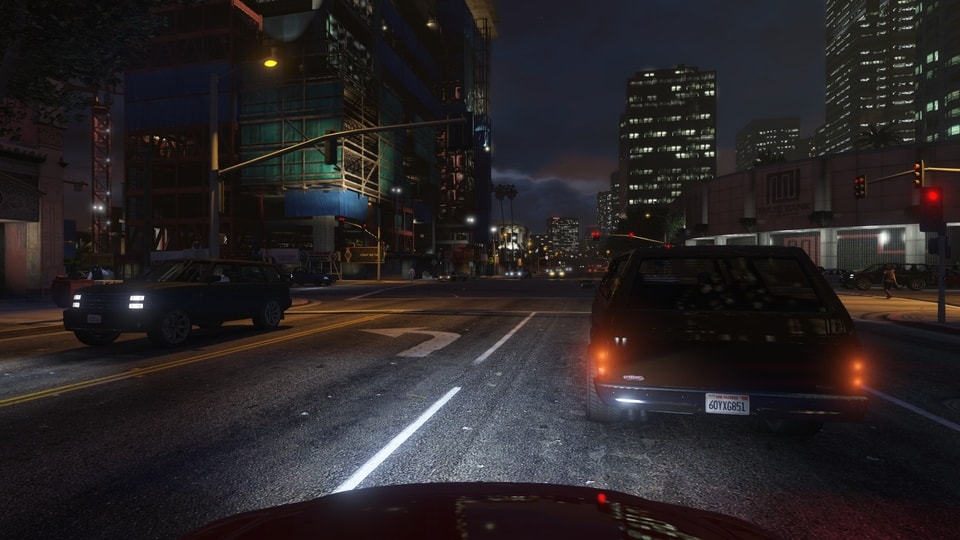} &
        \includegraphics[width=0.49\linewidth, keepaspectratio]{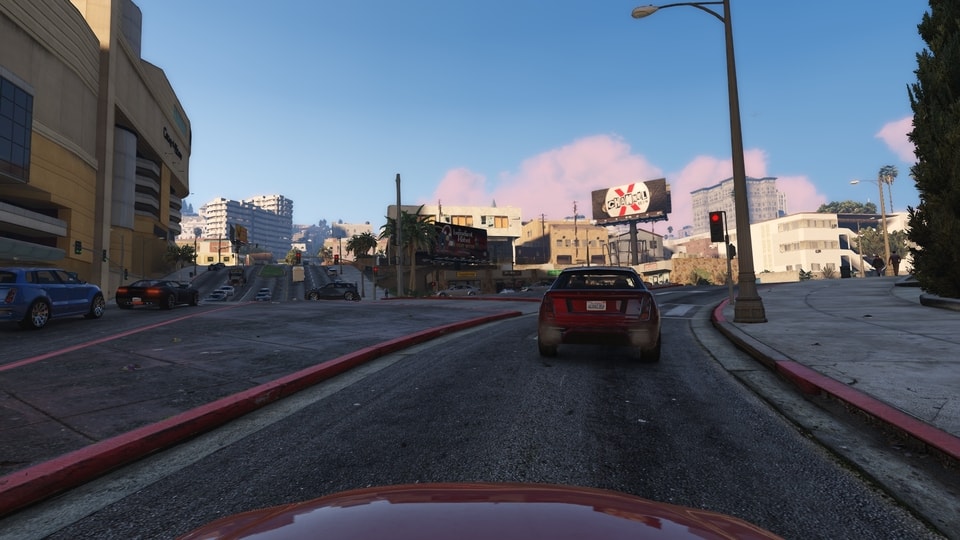}
    \end{tabular}
    }
	\caption{Examples of our dataset.}
	\label{fig:dataset_example}
\end{figure*}

\minisection{Examples of Our Dataset.}
\figref{fig:dataset_example} shows some visual examples in different time, weather and location.

\section{Training Details}
\label{apn:training_detail}

We apply different training strategies to our model to optimize the performance on different datasets.

\minisection{Training Procedure.}
We train on $4$ GPUs (effective mini-batch size is $20$). 
To keep the original image resolution can be divided by $32$, we use $1920\times1080$ resolution for GTA, $1920\times1216$ for Argoverse, and $1248\times384$ for KITTI. 
The anchors of RPN in Faster R-CNN span $4$ scales and $3$ ratios, in order to detect the small objects at a distance.
For GTA, we estimate both 2D detection and 3D center on the Faster RCNN pr-trained on Pascal VOC dataset.
For KITTI, we use RRC~\cite{Ren17RRC} detection results following a state-of-the-art method, BeyondPixels~\cite{MOTBeyondPixels}.
For Argoverse, we use Faster RCNN detection results of mmdetection~\cite{mmdetection} implementation pre-trained on COCO~\cite{lin2014mscoco} dataset.
Since both the results do not come with projected 3D centers, we estimate the center $c$ from the network in~\secref{sec:3d_estimation} of the main paper.

\section{Experiments}
\label{apn:experiments}

\minisection{Visual Odometry vs Ground Truth.}
To analyze the impact of ego-motion on tracking, we compare tracking results based on poses from GT camera and Mono-VO approach in \tabref{tab:ego_motion_vo_gt}.
The proposed method maintains similar MOTA using estimated camera pose.
Note that 3D tracking is compensated with GT ego-motion in the text.

\begin{table}[htpb]
    \ra{1.0}
    \centering
    \adjustbox{width=0.8\linewidth}{
        \begin{tabular}{cc|ccc}
        \toprule
        DATA & POSE & MOTA$\uparrow$ & MOTP$\uparrow$ & MM$\downarrow$ \\ 
        \midrule
        \multirow{2}{*}{KITTI} & VO & 71.079 & \textbf{88.233} & 0.784 \\
         & GT & \textbf{71.215} & 88.210 & \textbf{0.642} \\
        \bottomrule
    \end{tabular}
    }
    \caption{
    Ablation study of tracking performance with estimated (VO) or ground truth (GT) camera pose in KITTI training set. We find that VO can lead to similar results as GT.
    }
    \label{tab:ego_motion_vo_gt}
\end{table}

\minisection{Inference Time.}
The total inference time is $203$ ms on a single P100 GPU (see Table~\ref{tab:inference_time} for details). 
Note that the inference time on KITTI benchmark focus only on the non-detection part ($92$ ms).

\begin{table}[htpb]
\begin{tabular}{c|ccc}
\toprule
 & Detection & 3D Estimation & Tracking\\ \midrule
Run Time & 0.111 & 0.062 & 0.030 \\
\bottomrule
\end{tabular}
\caption{Inference Time (second) of our proposed framework on KITTI tracking benchmark. Note that elapsed time for object detection is not included in the specified runtime of KITTI benchmark.}
\label{tab:inference_time}
\end{table}

\minisection{Tracking policy.}
In the association, we keep all tracklets until they disappear from our tracking range (e.g. $−10$m to $100$m) or die out after $20$ time-steps. 
\tabref{tab:tracking_range} shows the ablation comparison of using location and max-age.
Since the max-age affects the length of each tracklet, we prefer to keep the tracklet a longer period (\ie $>15$) to re-identify possible detections using 3D location information.

\begin{table}[thbp]
\caption{Comparison of performance using different location and maxage in our tracking pipeline.
The hyphen - represents not using the information in tracking.}
\adjustbox{width=\linewidth}{
\begin{tabular}{c|c|cccc|cc}
\toprule
Max-Age & Location & IDS$\downarrow$ & FRAG$\downarrow$ & FP$\downarrow$ & FN$\downarrow$ & MOTA$\uparrow$ & MOTP$\uparrow$\\
\midrule
30     &      -     & 513 & 710 & 341 & 1571 & 89.93 & 90.40 \\
20     &      -    & 373 & 571 & 341 & 1571 & 90.51 & 90.40 \\
-      & \checkmark & 316 & 513 & 341 & 1571 & 90.74 & 90.40 \\ 
30     & \checkmark & 290 & 488 & 341 & 1571 & 90.85 & 90.40 \\ 
20     & \checkmark & \textbf{261} & \textbf{459} & \textbf{341} & \textbf{1571} & \textbf{90.97} & \textbf{90.40} \\
\bottomrule
\end{tabular}
}
\label{tab:tracking_range}
\end{table}

\minisection{Data Association Weights.}
During our experiment, we use different weights of appearance, 2D overlap, and 3D overlap for corresponding methods.
We select weight ratios based on the results of our validation set.

For None and 2D, we use $w_{\subfix{app}}=0.0$,$w_{\subfix{2D}}=1.0$, $w_{\subfix{3D}}=0.0$.
For 3D related methods, we give $w_{\subfix{3D}}=1.0$ and $w_{\subfix{2D}}=0.0$.
For Deep related methods, we times $w_{\subfix{2D}}$ or $w_{\subfix{3D}}$ with $0.7$ and $0.3$ for $w_{\subfix{app}}$.

\minisection{Tracking Performance on KITTI dataset.}
As mentioned in the main paper, we resort to Argoverse~\cite{Chang2019argoverse} and KITTI~\cite{geiger2012we} tracking benchmarks to compare our model abilities in the real-world scenario.
Our monocular 3D tracking method outperforms all the published methods upon the time of submission.
Results are listed in \tabref{tab:kitti_full_tracking}.

\begin{table}[htpb]
\caption{Tracking performance on the testing set of KITTI tracking benchmark.
Only published methods are reported.
$\dagger$ sign means 3D information is used.
}
\centering
\adjustbox{width=\linewidth}{
    \begin{tabular}{lrrrrrr}
        \toprule
        {Benchmark} & {MOTA} $\uparrow$& {MOTP} $\uparrow$ & {MT} $\uparrow$& {ML} $\downarrow$ & {FP} $\downarrow$ & {FN} $\downarrow$ \\
        \midrule
        Ours & \textbf{84.52} & 85.64 & \textbf{73.38} & \textbf{2.77} & 705 & \textbf{4242} \\
        BeyondPixels$^\dagger$~\cite{MOTBeyondPixels} & 84.24 & \textbf{85.73} & 73.23 & \textbf{2.77} & 705 & 4247 \\
        PMBM$^\dagger$~\cite{Scheidegger2018} & 80.39 & 81.26 & 62.77 & 6.15 & 1007 & 5616 \\
        extraCK~\cite{gunduz2018lightweight} & 79.99 & 82.46 & 62.15 & 5.54 & 642 & 5896 \\
        MCMOT-CPD~\cite{Lee2016MCMOTCPD} & 78.90 & 82.13 & 52.31 & 11.69 & \textbf{316} & 6713 \\
        NOMT*~\cite{choi2015nomt} & 78.15 & 79.46 & 57.23 & 13.23 & 1061 & 6421 \\
        MDP~\cite{xiang2015mdptrack} & 76.59 & 82.10 & 52.15 & 13.38 & 606 & 7315 \\
        DSM~\cite{frossard2018dsm} & 76.15 & 83.42 & 60.00 & 8.31 & 578 & 7328 \\
        SCEA~\cite{hong2016scea} & 75.58 & 79.39 & 53.08 & 11.54 & 1306 & 6989 \\
        CIWT~\cite{Osep17ICRAciwt} & 75.39 & 79.25 & 49.85 & 10.31 & 954 & 7345 \\
        NOMT-HM*~\cite{choi2015nomt} & 75.20 & 80.02 & 50.00 & 13.54 & 1143 & 7280 \\
        mbodSSP~\cite{lenz2015followme} & 72.69 & 78.75 & 48.77 & 8.77 & 1918 & 7360 \\
        \bottomrule
    \end{tabular}
}
\label{tab:kitti_full_tracking}
\end{table}

\minisection{Ablation Study.}
The ablation study of tracking performance can be found in \tabref{tab:gta_tracking_full}.
From the ablation study, we observe the huge performance improvement, especially in track recall (TR) for a $8\%$ and Accuracy (MOTA) for a $18\%$ relative improvement, from 2D to 3D, which indicates 3D information helps to recover unmatched pairs of objects.
We implement our 2D baselines deriving from Sort~\cite{Bewley2016sort}, Deep Sort~\cite{Wojke2017deepsort} tracker for inter-frame ROIs matching in 2D.
Modeling motion in 3D coordinate is more robust in linking candidates than 2D motion. 
Moreover, adding deep feature distinguishes two near-overlapping objects, our false negative (FN) rate drops with an observable margin. 
With depth-order matching and occlusion-aware association, our model filters out $6-8\%$ possible mismatching trajectories. 

\begin{table*}[htpb]
    \ra{1.2}
    \adjustbox{width=\linewidth}{
        \begin{tabular}{ccc|ccccccccc}
        \toprule
        Motion & Deep & Order & MOTA $\uparrow$ & MOTP $\uparrow$ & TP $\uparrow$ & TR $\uparrow$ & MM $\downarrow$ & NM $\downarrow$ & RM $\downarrow$ & FP $\downarrow$ & FN $\downarrow$ \\ 
        \midrule
        - & - & - & 5.056 & 67.726 & 79.445 & 71.102 & 6.31 & 1.636 & 5.839 & 25.997 & 62.637 \\
        2D~\cite{Bewley2016sort} & - & - & 57.042 & 82.794 & 86.905 & 77.058 & 3.739 & 1.064 & 3.418 & 6.085 & 33.134 \\
        2D$^*$~\cite{Wojke2017deepsort} & \checkmark & - & 65.892 & 81.86 & 90.607 & 87.186 & 4.796 & 4.096 & 3.898 & 10.099 & 19.213  \\
        3D & - & - & 69.616 & 84.776 & 85.947 & 84.202 & 1.435 & 0.511 & 1.283 & 7.651 & 21.298\\
        3D & \checkmark & - & 69.843 & 84.523 & 90.242 & 87.113 & 2.269 & 1.798 & 1.672 & 9.341 & 18.547 \\
        3D & - & \checkmark & 70.061 & \textbf{84.845} & 85.952 & 84.377 & \textbf{1.317} & \textbf{0.403} & \textbf{1.184} & \textbf{7.303} & 21.319 \\
        3D & \checkmark & \checkmark & 70.126 & 84.494 & 90.155 & \textbf{87.432} & 2.123 & 1.717 & 1.512 & 9.574 & 18.177 \\
        LSTM & \checkmark & \checkmark & \textbf{70.439} & 84.488 & \textbf{90.651} & 87.374 & 1.959 & 1.547 & 1.37 & 9.508 & \textbf{18.094} \\ 
        \midrule
        - & - & - & 4.609 & 71.064 & 74.941 & 83.594 & 14.535 & 3.735 & 13.834 & 31.326 & 49.531 \\
        2D~\cite{Bewley2016sort} & - & - & 42.748 & 81.878 & \textbf{70.003} & 84.292 & 9.172 & 2.077 & 8.755 & \textbf{16.731} & 31.350 \\
        2D$^*$~\cite{Wojke2017deepsort} & \checkmark & - &  48.518 & 81.479 & 66.381 & 88.222 & 7.270 & 2.683 & 6.738 & 21.223 & 22.989 \\
        3D & - & - & 54.324 & 83.914 & 64.986 & 90.869 & 3.032 & 0.799 & 2.820 & 23.574 & 19.070 \\
        3D & \checkmark & - & 54.855 & 83.812 & 68.235 & 91.687 & 2.087 & 0.776 & 1.864 & 25.223 & 17.835\\
        3D & - & \checkmark & 54.738 & 84.045 & 64.001 & 90.623 & 3.242 & \textbf{0.533} & 3.102 & 22.120 & 19.899 \\
        3D & \checkmark & \checkmark & \textbf{55.218} & 83.751 & 68.628 & \textbf{92.132} & \textbf{1.902} & 0.723 & \textbf{1.688} & 25.578 & \textbf{17.302} \\
        LSTM & \checkmark & \checkmark & 55.150 & \textbf{83.780} & 69.860 & 92.040 & 2.150 & 0.800 & 1.920 & 25.220 & 17.470 \\
        \bottomrule
        \end{tabular}
        }
        \caption{
        Ablation study of tracking performance with different methods and features in GTA dataset.
        The upper half of the table shows evaluation of targets within $100$ meters while the lower within $150$ meters.
        Modeling motion in 3D coordinate with the help of our depth estimation is more robust in linking candidate tracklets than 2D motion. 
        Using deep feature in correlation can reduce the false negative (FN) rate. 
        Using depth-order matching and occlusion-aware association filter out relatively $6-8\%$ mismatching trajectories.  
        Replacing 3D kalman filter with LSTM module to re-estimate the location using temporal information.
        }
        \label{tab:gta_tracking_full}
\end{table*}

\begin{figure*}[htpb]
    \minipage{0.32\textwidth}
        \includegraphics[width=1.0\linewidth, keepaspectratio]{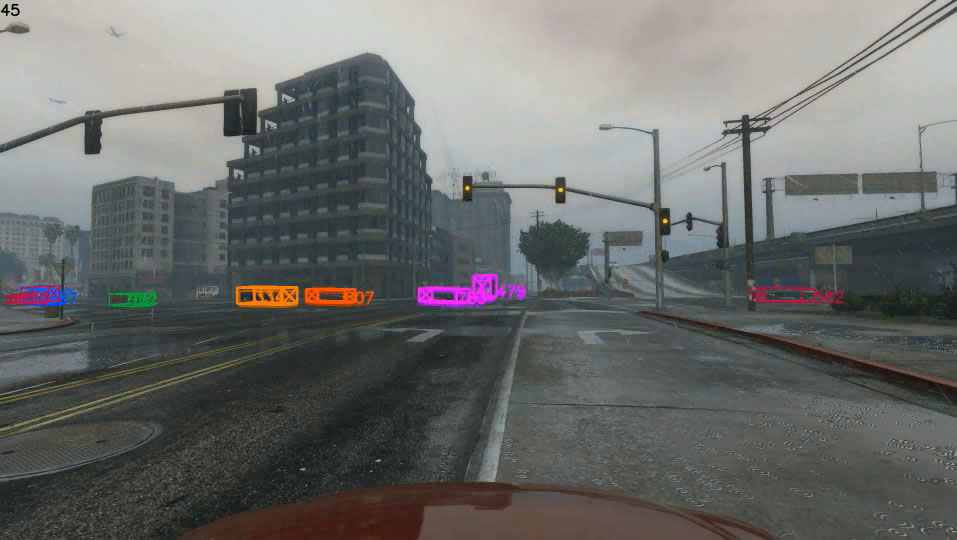}
        \includegraphics[width=1.0\linewidth,  keepaspectratio]{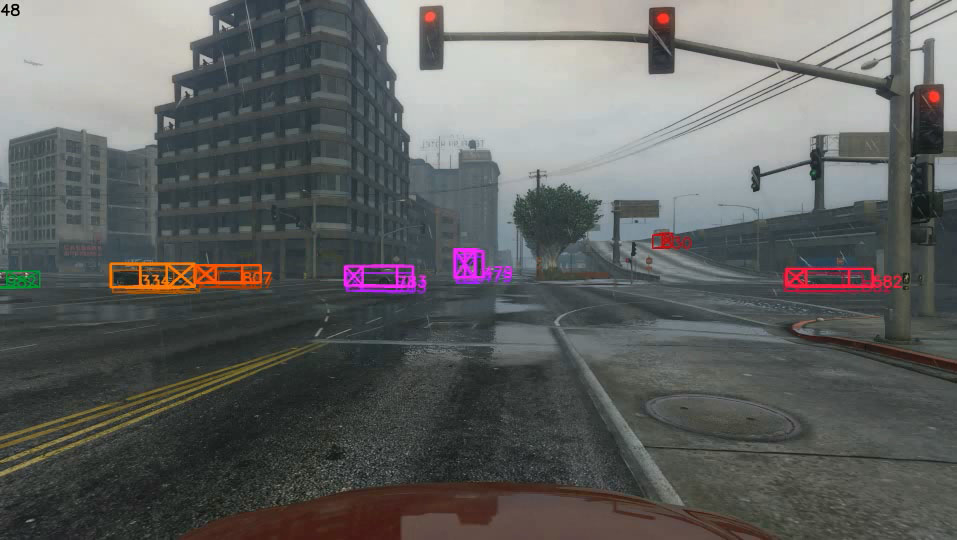}
        \includegraphics[width=1.0\linewidth, keepaspectratio]{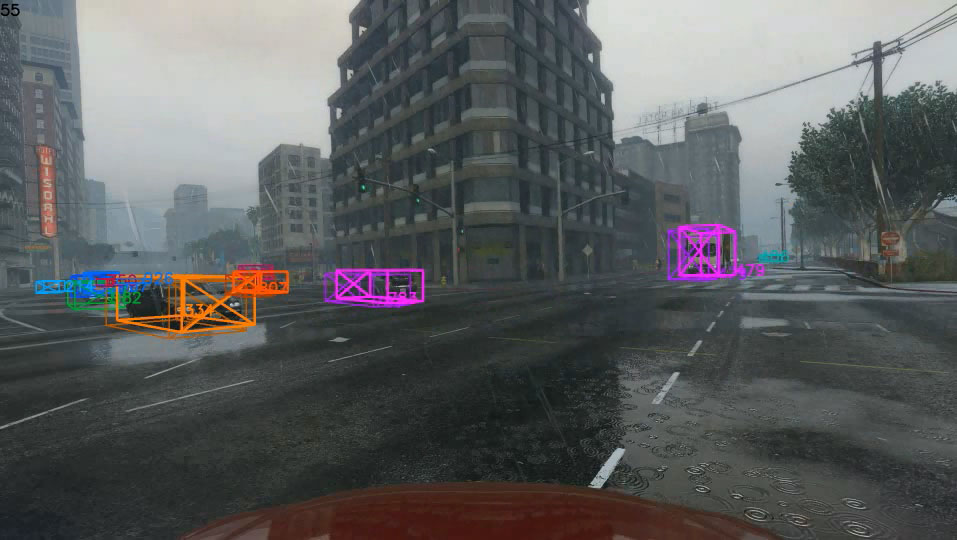}
        \includegraphics[width=1.0\linewidth,  keepaspectratio]{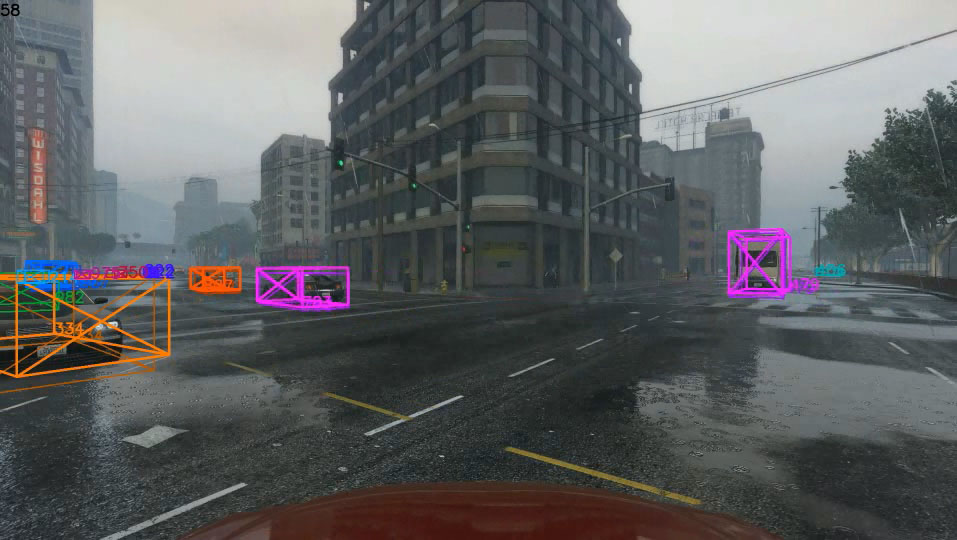}
        \includegraphics[width=1.0\linewidth,  keepaspectratio]{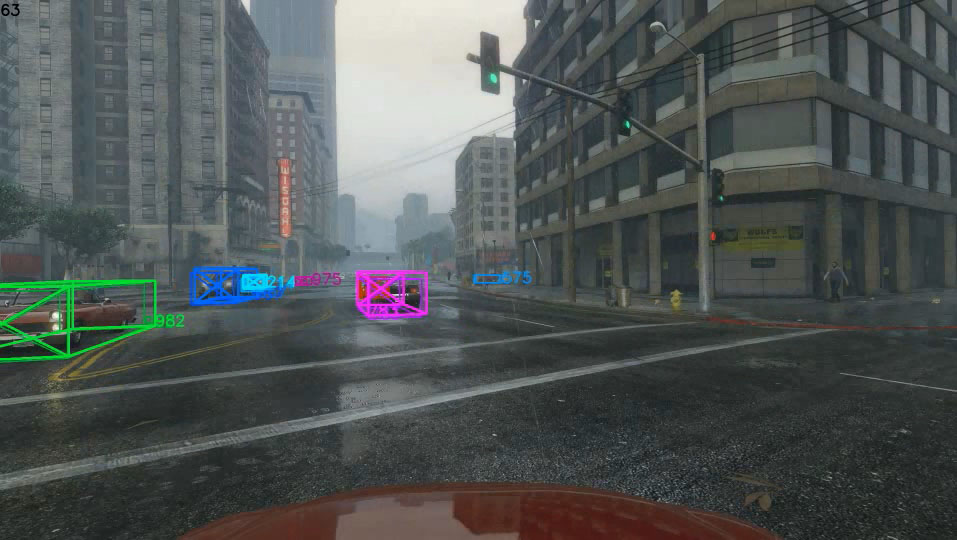}
    \endminipage
    \minipage{0.18\textwidth}
        \includegraphics[width=1.0\linewidth, keepaspectratio]{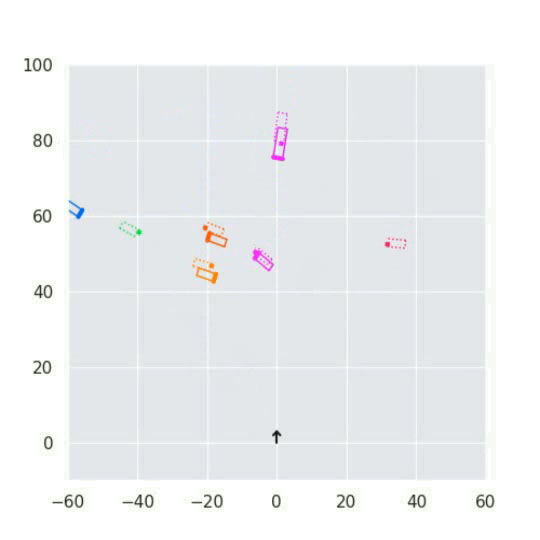}
        \includegraphics[width=1.0\linewidth,  keepaspectratio]{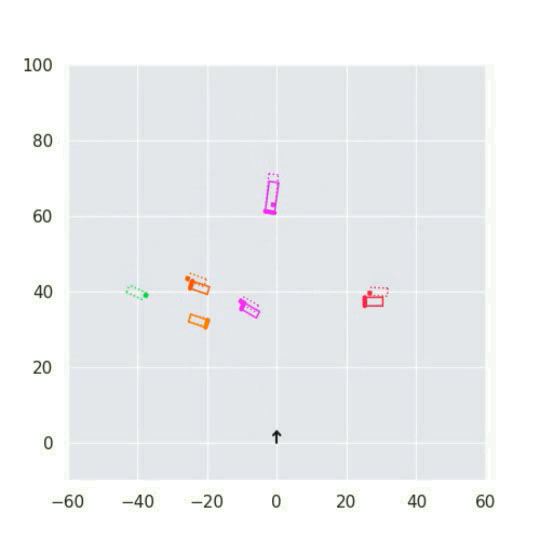}
        \includegraphics[width=1.0\linewidth, keepaspectratio]{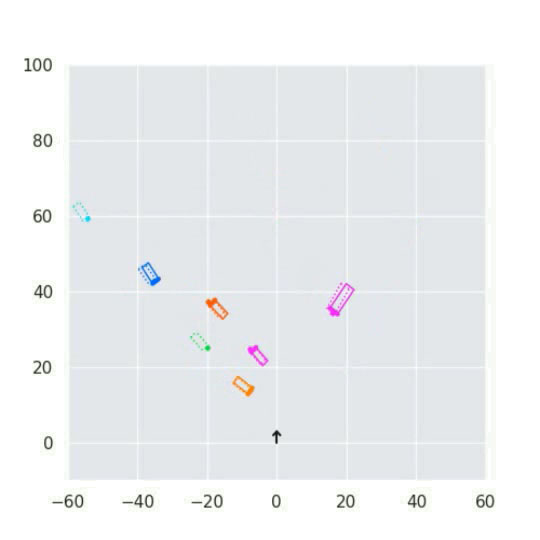}
        \includegraphics[width=1.0\linewidth,  keepaspectratio]{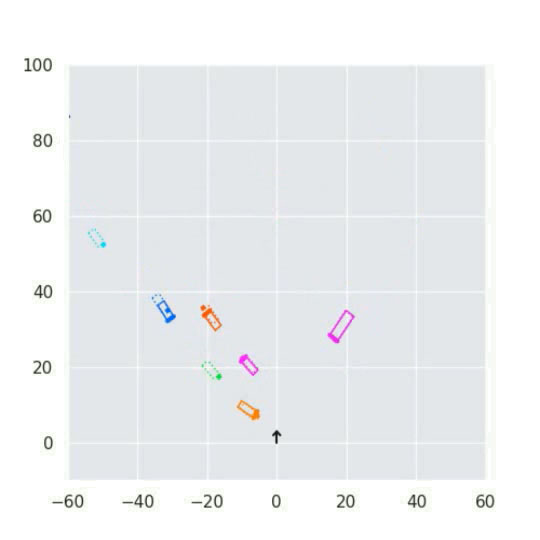}
        \includegraphics[width=1.0\linewidth,  keepaspectratio]{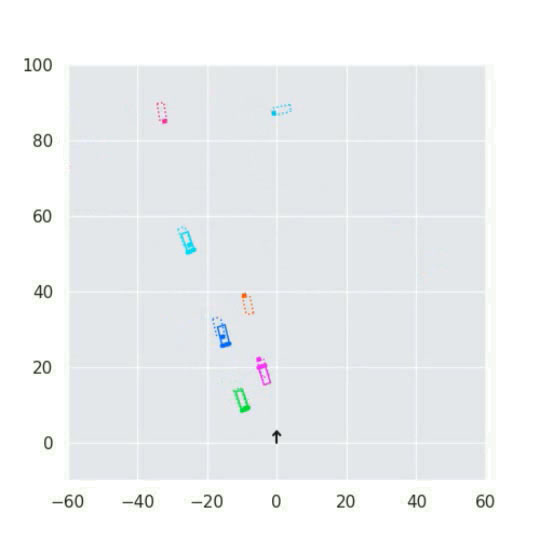}
    \endminipage
    \hfill
    \minipage{0.32\textwidth}
        \includegraphics[width=1.0\linewidth, keepaspectratio]{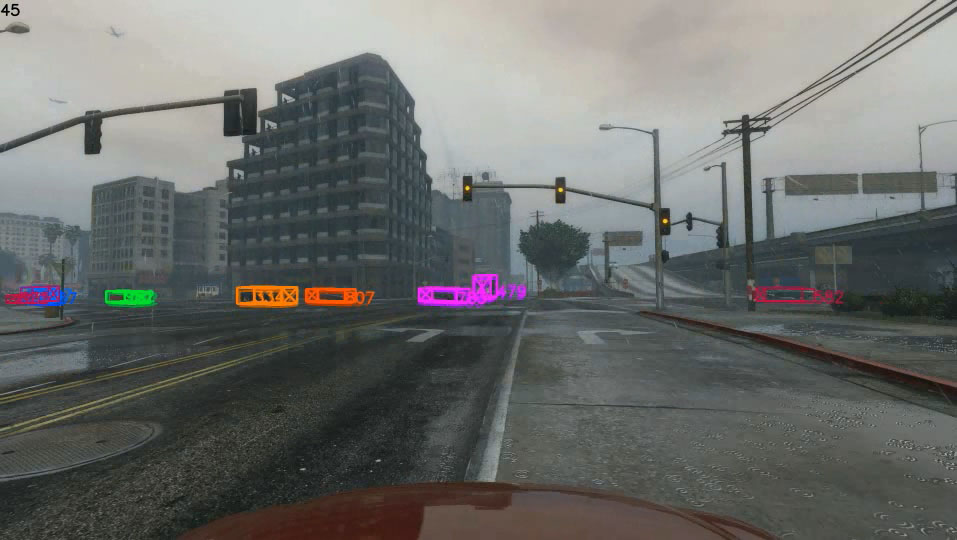}
        \includegraphics[width=1.0\linewidth,  keepaspectratio]{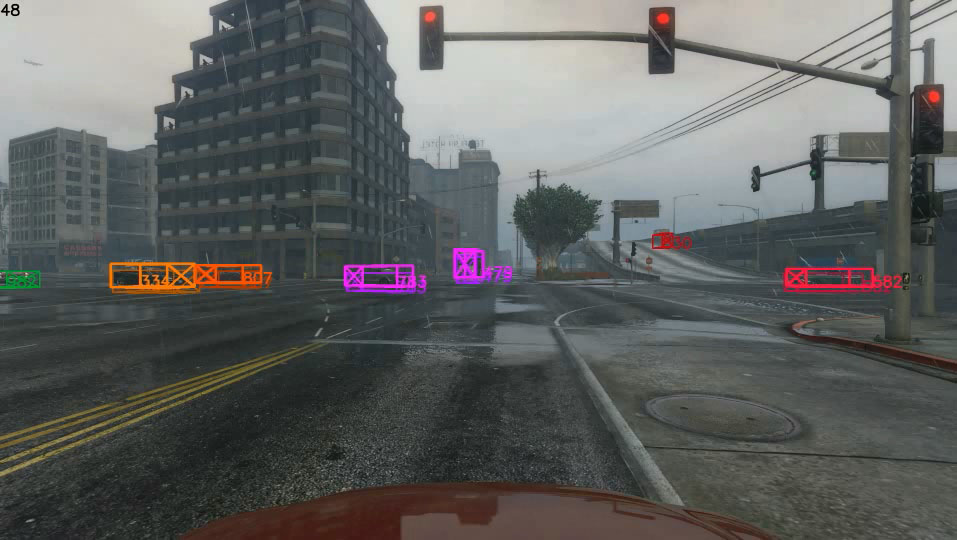}
        \includegraphics[width=1.0\linewidth, keepaspectratio]{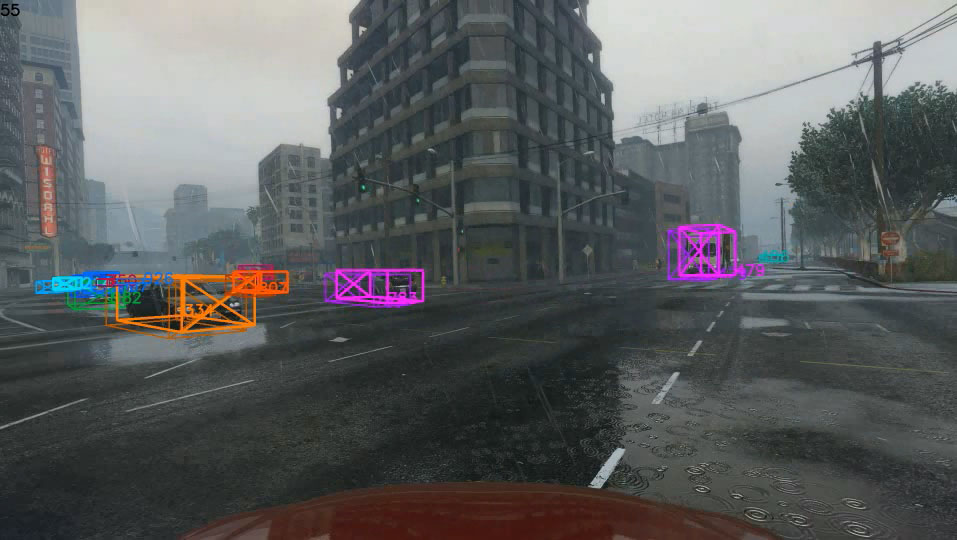}
        \includegraphics[width=1.0\linewidth,  keepaspectratio]{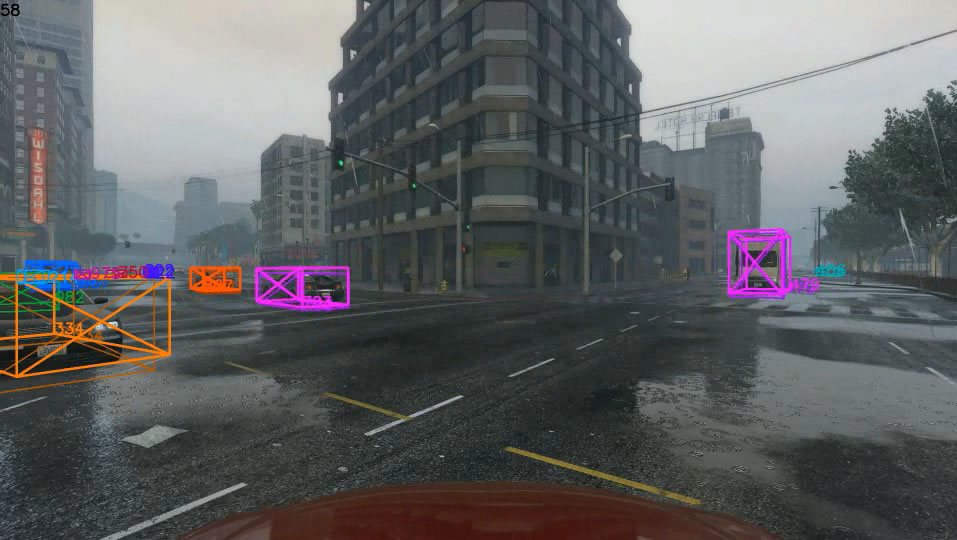}
        \includegraphics[width=1.0\linewidth,  keepaspectratio]{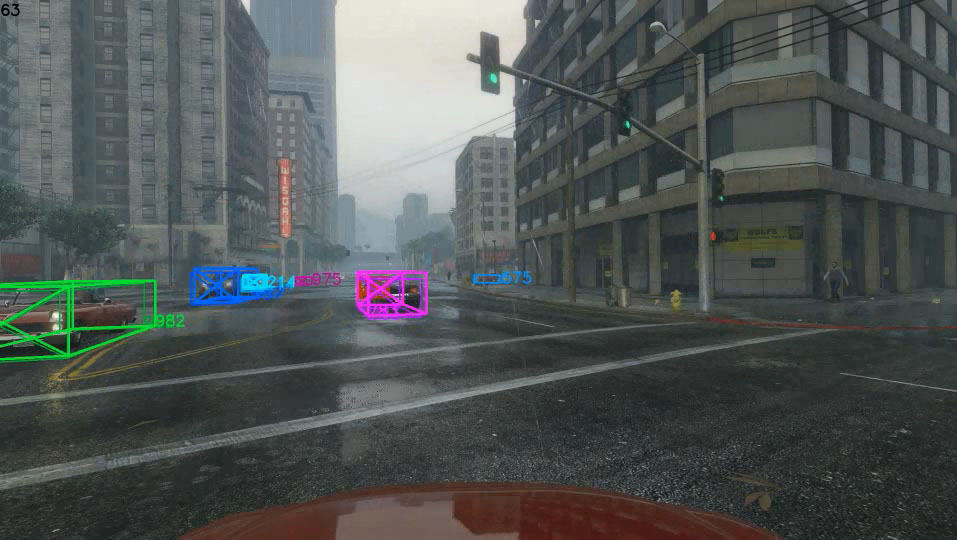}
    \endminipage
    \minipage{0.18\textwidth}
        \includegraphics[width=1.0\linewidth, keepaspectratio]{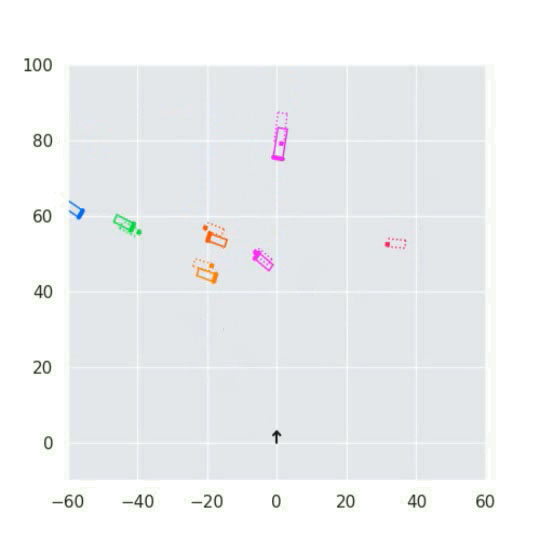}
        \includegraphics[width=1.0\linewidth,  keepaspectratio]{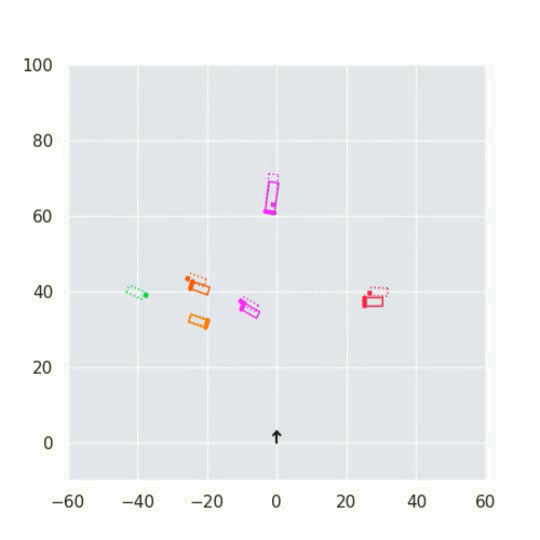}
        \includegraphics[width=1.0\linewidth, keepaspectratio]{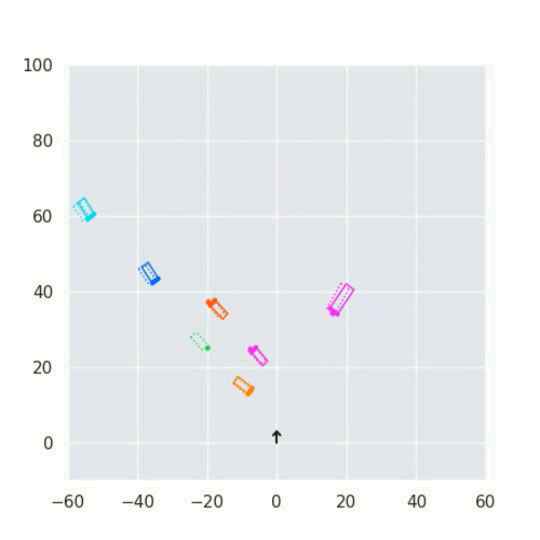}
        \includegraphics[width=1.0\linewidth,  keepaspectratio]{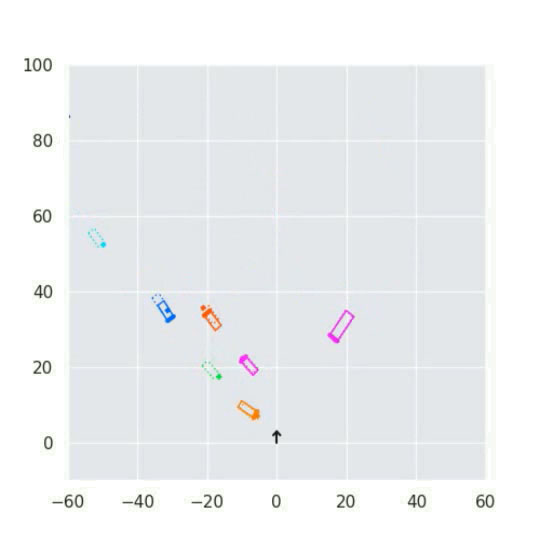}
        \includegraphics[width=1.0\linewidth,  keepaspectratio]{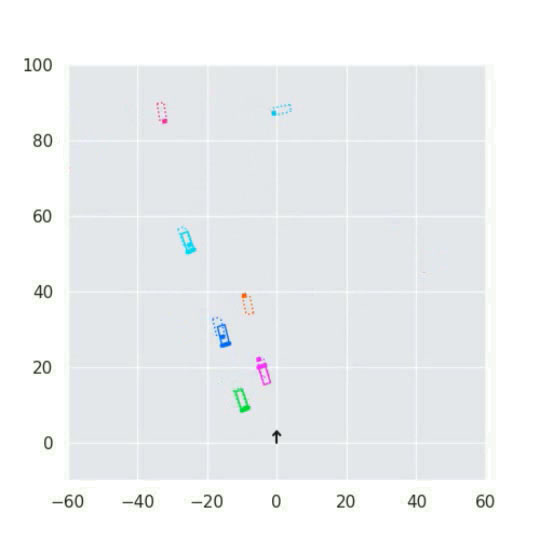}
    \endminipage
	\caption{Qualitative comparison results of 3D Kalman and our method.}
	\label{fig:qualitative_gta}
\end{figure*}
\begin{figure*}[htpb]
    \minipage{0.49\textwidth}
        \includegraphics[width=1.0\linewidth, keepaspectratio]{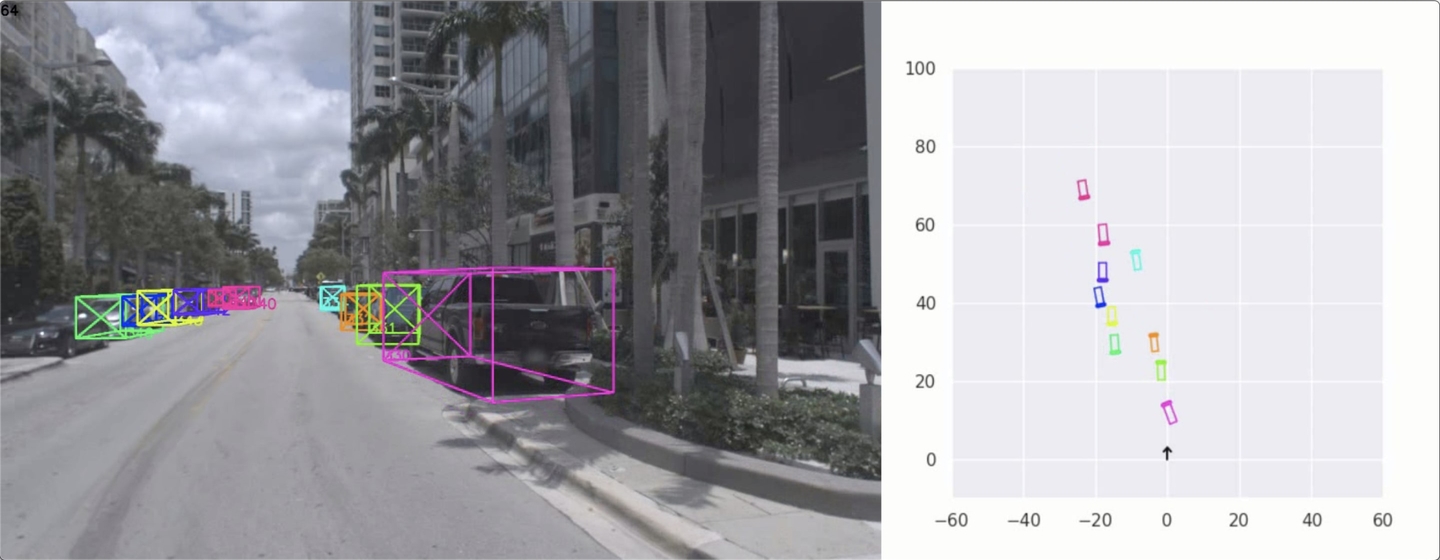}
        \includegraphics[width=1.0\linewidth,  keepaspectratio]{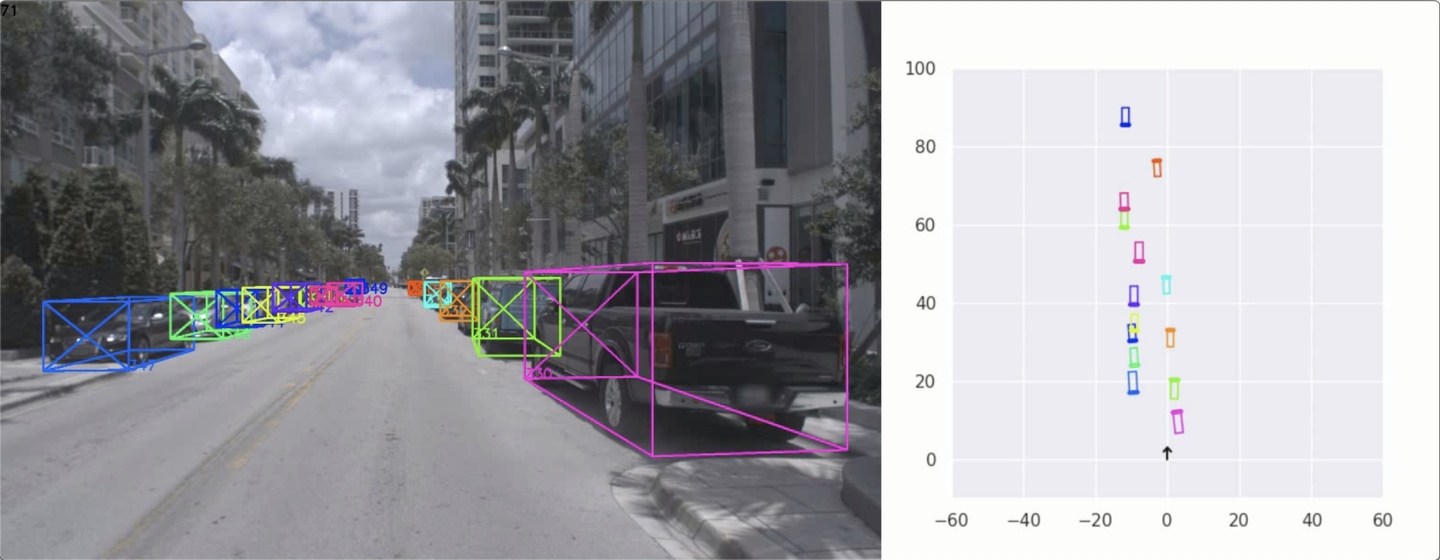}
        \includegraphics[width=1.0\linewidth, keepaspectratio]{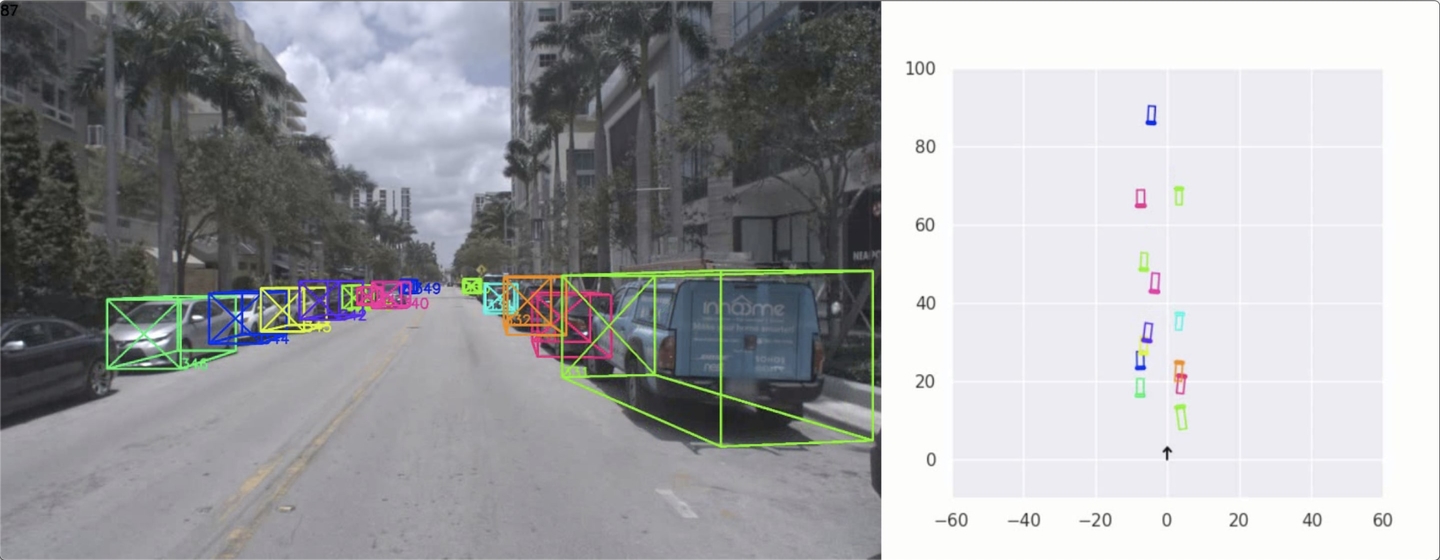}
        \includegraphics[width=1.0\linewidth,  keepaspectratio]{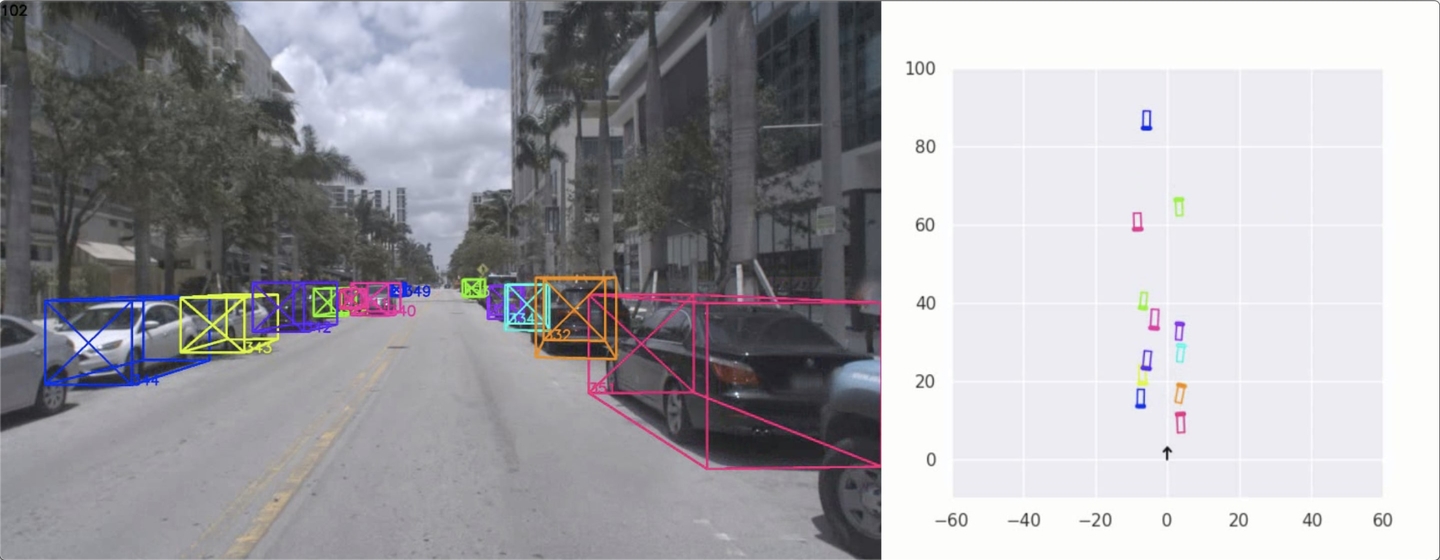}
        \includegraphics[width=1.0\linewidth,  keepaspectratio]{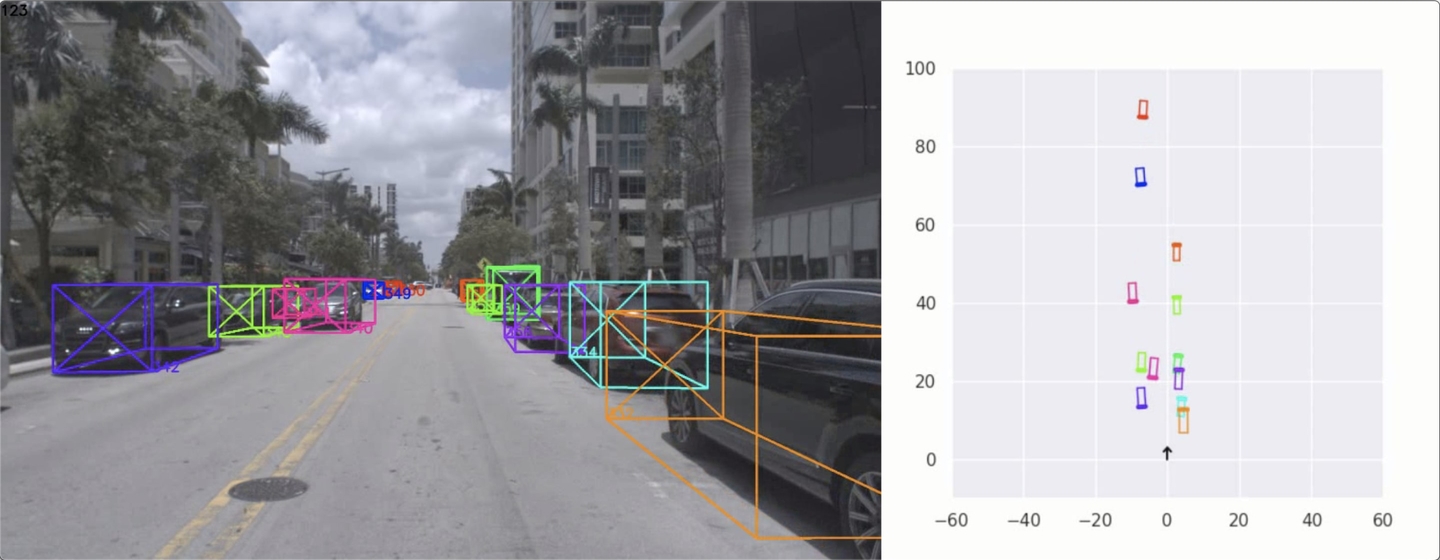}
    \endminipage
    \minipage{0.49\textwidth}
        \includegraphics[width=1.0\linewidth, keepaspectratio]{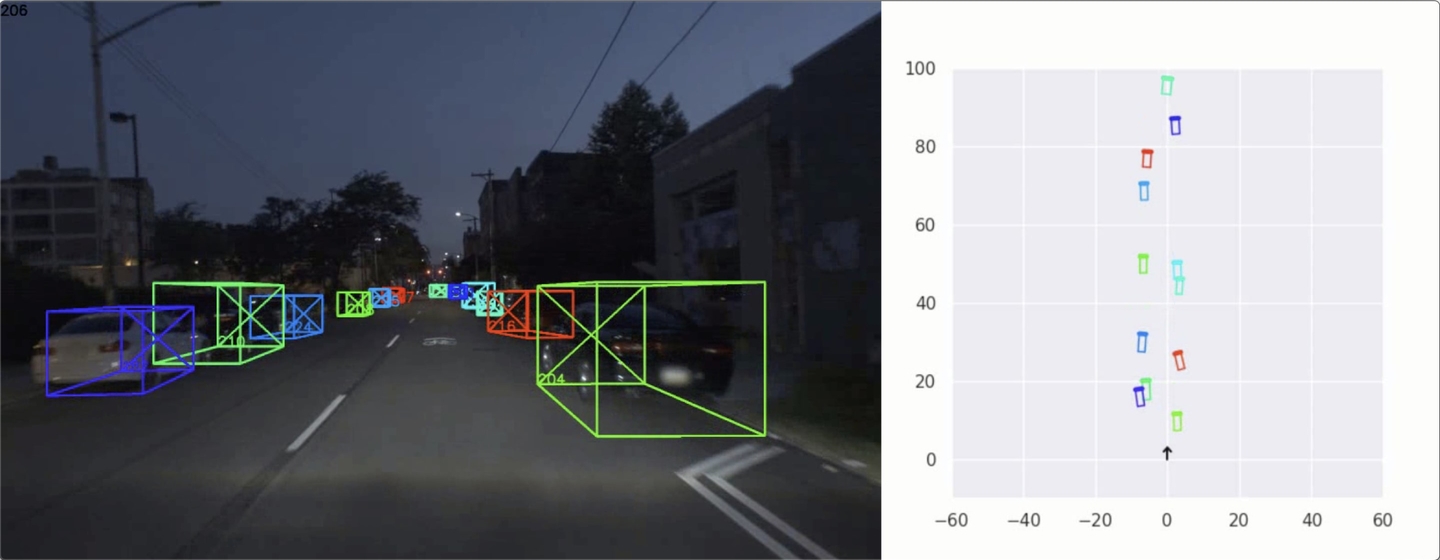}
        \includegraphics[width=1.0\linewidth,  keepaspectratio]{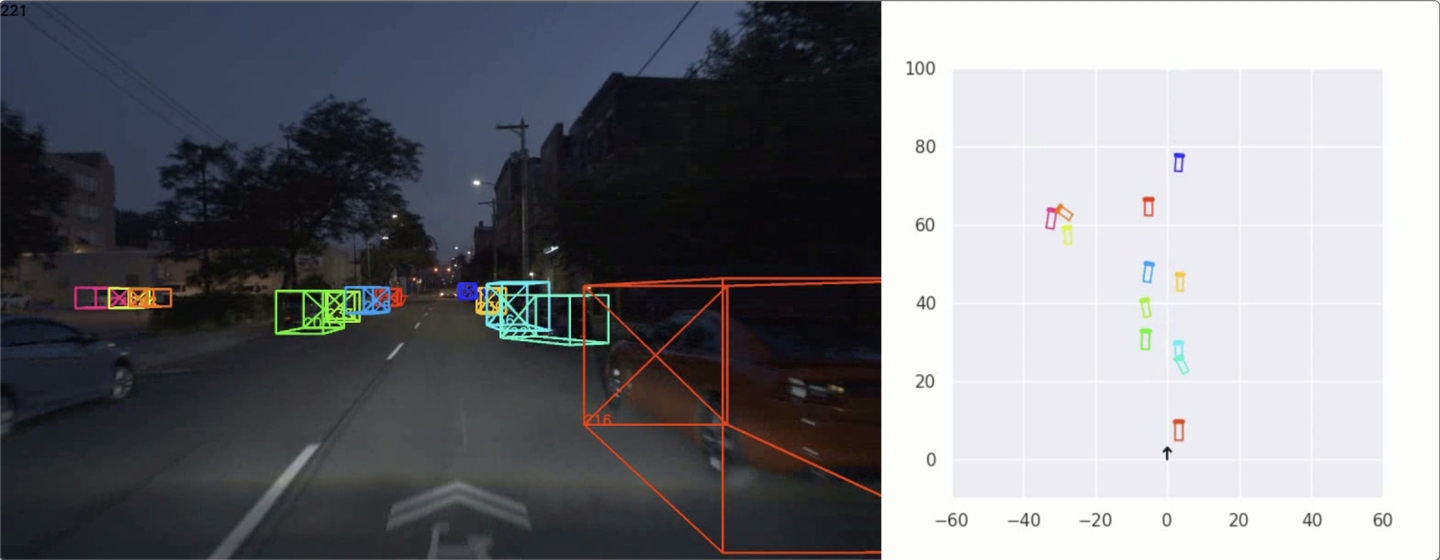}
        \includegraphics[width=1.0\linewidth, keepaspectratio]{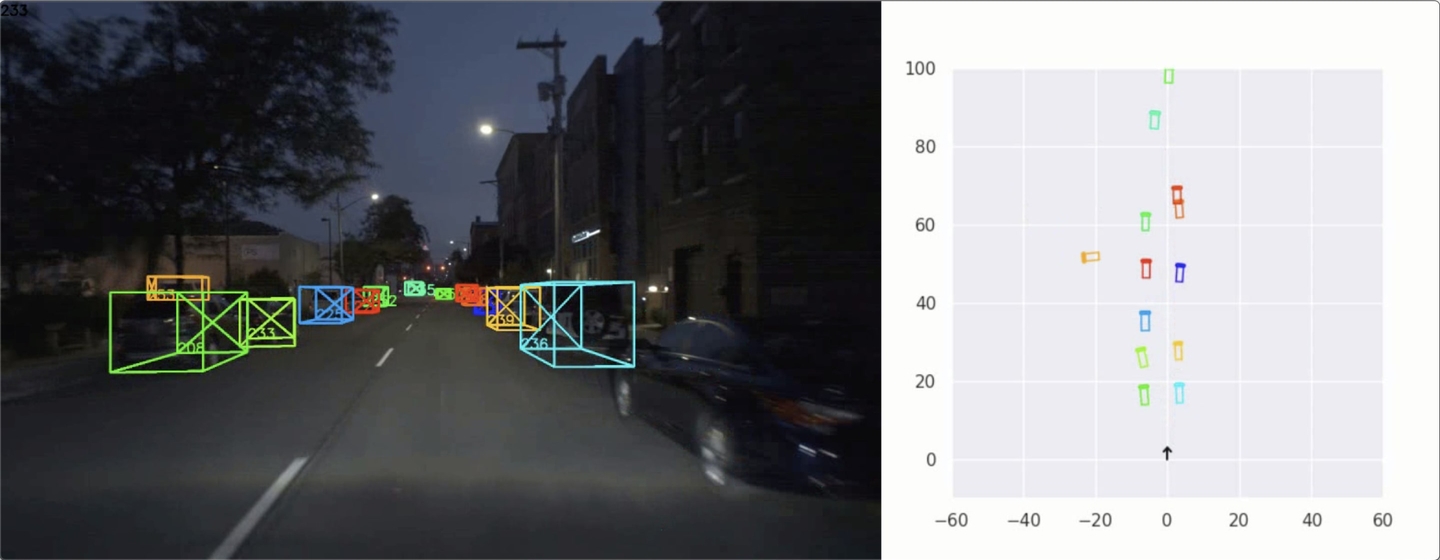}
        \includegraphics[width=1.0\linewidth,  keepaspectratio]{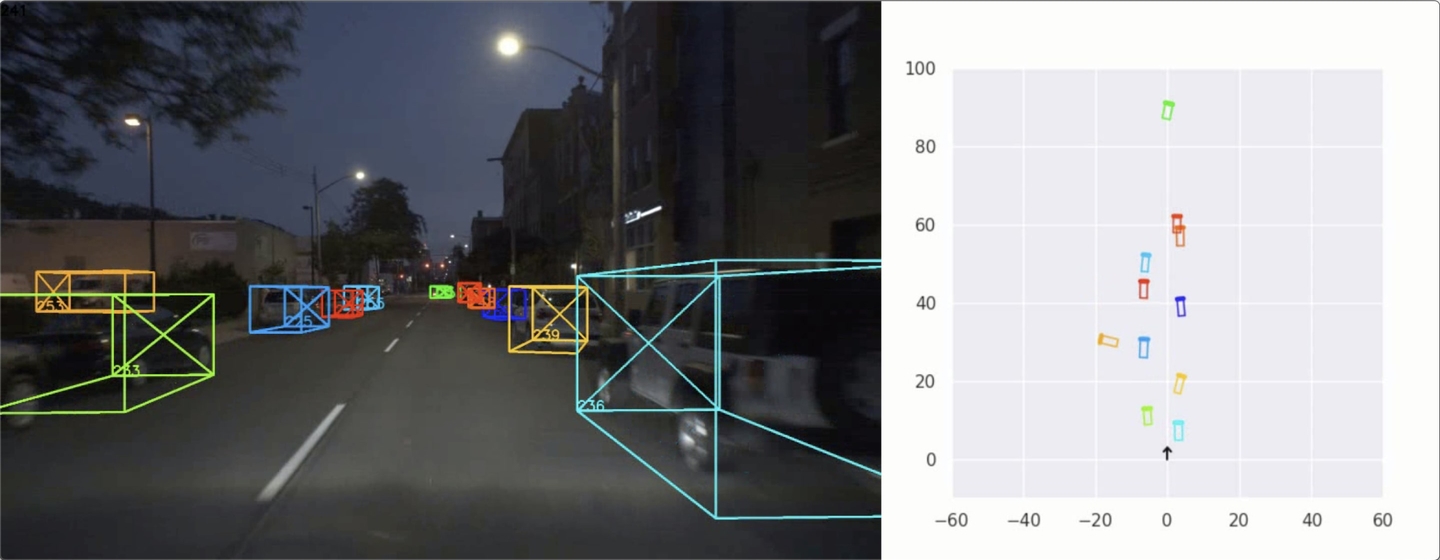}
        \includegraphics[width=1.0\linewidth,  keepaspectratio]{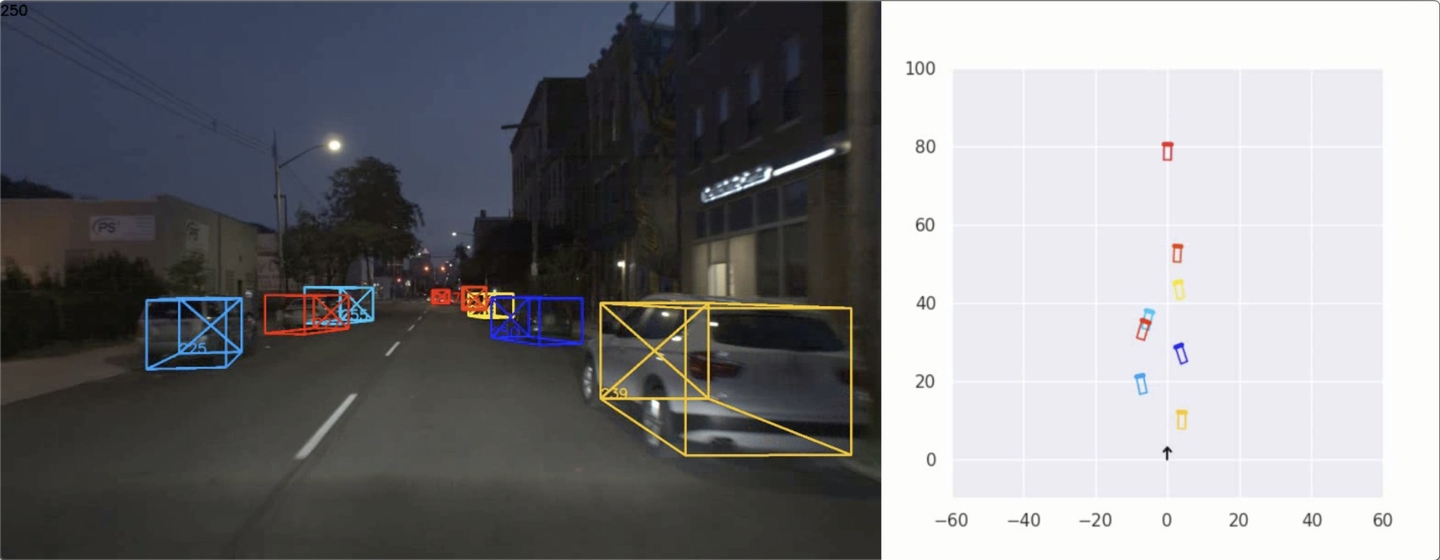}
    \endminipage
	\caption{Qualitative results of our method on Argoverse.}
	\label{fig:qualitative_argo}
\end{figure*}

\minisection{Qualitative results.}
We show our evaluation results in \figref{fig:qualitative_gta} on our GTA test set and in \figref{fig:qualitative_argo} on Argoverse validation set.
The method comparison on GTA shows that our proposed 3D Tracking outperforms the strong baseline which using 3D Kalman Filter.
The light dashed rectangular represents the ground truth vehicle while the solid rectangular stands for the predicted vehicle in bird's eye view.
The qualitative results on Argoverse demonstrate our monocular 3D tracking ability under day and night road conditions without the help of LiDAR or HD maps.
The figures are best viewed in color.

\minisection{Evaluation Video.}
We have uploaded a demo video which summaries our main concepts and demonstrates video inputs with estimated bounding boxes and tracked trajectories in bird's eye view. We show the comparison of our method with baselines in both our GTA dataset and real-world videos. 

